\documentclass[lettersize,journal]{IEEEtran}
\usepackage{amsmath,amsfonts}
\usepackage{array}
\usepackage[caption=false,font=normalsize,labelfont=sf,textfont=sf]{subfig}
\usepackage{textcomp}
\usepackage{stfloats}
\usepackage{url}
\usepackage{verbatim}
\usepackage{graphicx}
\usepackage[table]{xcolor}
\usepackage{cite}
\usepackage{float} 
\usepackage{booktabs}
\usepackage{cuted}
\usepackage{caption}
\usepackage{tabularx}
\usepackage{stackengine}
\begin{document}

\definecolor{bruno}{rgb}{0.0, 0.5, 0.0}
\definecolor{hermes}{rgb}{0.07, 0.04, 0.56}
\definecolor{ane}{rgb}{1.0, 0.44, 0.37}
\definecolor{ali}{rgb}{0,0.502,1}
\newcommand{\ane}[1]{\textcolor{ane}{#1}}
\newcommand{\ali}[1]{\textcolor{ali}{#1}}

\title{WiFuse: An Attention Mechanism for Human Activity Recognition using Fused CSI Amplitude and Delay–Doppler Channel Features}
\author{Alison M. Fernandes,~Hermes I. Del Monego,~Bruno S. Chang, Anelise Munaretto,\\Hélder M. Fontes and Rui L. Campos%
\thanks{A. M. Fernandes, H. I. Del Monego, B. S. Chang, and A. Munaretto are with the Universidade Tecnológica Federal do Paraná (UTFPR), Programa de Pós-Graduação em Engenharia Elétrica e Informática Industrial (CPGEI-CT), Av. Sete de Setembro, 3165, Curitiba, 80230-901, Brazil. Hélder M. Fontes and Rui L. Campos are with INESC TEC, Faculdade de Engenharia, Universidade do Porto, Rua Dr. Roberto Frias, Porto, 4200-465, Porto, Portugal. Part of this work was submitted for presentation in IEEE ISAC 2026.}%
\thanks{Corresponding author: Bruno S. Chang (e-mail: bschang@utfpr.edu.br).}%
\thanks{This work was supported by the Coordination for the Improvement of Higher Education Personnel (CAPES) under ROR identifier: 00x0ma614 for the Article Processing Charge. This work was also supported by CNPq (444629/2024-6). This work was also partially supported by Instituto de Engenharia de Sistemas e Computadores, Pesquisa e Desenvolvimento do Brasil (INESC P\&D Brasil). This research is part of the Instituto Nacional de Ciência e Tecnologia (INCT) of Intelligent Communications Networks and the Internet of Things (ICoNIoT), funded by CNPq (proc. 405940/2022-0) and CAPES (Finance Code 88887.954253/2024-00).}%
}

\markboth{Journal of \LaTeX\ Class Files,~Vol.~XX, No.~X, Month~202X}%
{Fernandes \MakeLowercase{\textit{et al.}}: Title of Your Paper}

\maketitle

\begin{abstract}

Recently, Wi-Fi sensing has played a significant role in Human Activity Recognition (HAR), as it enables the detection of various activities using only Wi-Fi signals, ensuring privacy and remaining non-intrusive for the user. However, environmental characteristics such as reflective surfaces, hardware offsets, and other physical impairments affect recognition by the neural network, subsequently causing errors and significantly reducing model accuracy. To overcome this problem we present the WiFuse framework, a dual-stream Channel State Information (CSI) framework for human activity recognition (HAR) that pairs denoised time-domain amplitude variations with 2D-FFT–derived Delay–Doppler motion representations computed from the sanitized channel phase. The fused representation feeds a hybrid ResNet–Temporal Convolutional Network (TCN) neural architecture augmented with channel and spatio-temporal attention, where the ResNet extracts spatial–spectral features and the TCN models long-range temporal dependencies; a decoupled two-stage transfer learning strategy is employed to improve optimization stability and feature reuse. We conduct extensive experiments on two public datasets, including comparisons against state-of-the-art methods and alternative hybrid architectures, ablation studies, and cross-dataset and domain-adaptation evaluations. The proposed framework reaches an overall accuracy of up to 95.28\% across the four environments of the XRF55 dataset and up to 98.20\% on the multi-user Wi-MIR dataset. Overall, the results indicate that combining amplitude and Delay–Doppler representations within a dual-stream strategy, enhanced by transfer learning, improves recognition performance under conditions that typically degrade deep neural networks, such as class overlap, multipath propagation, noise, and interference.
\end{abstract}


\section{Introduction}
\label{sec:introduction}

Nowadays, several studies are being conducted based on the principles of Wi-Fi sensing for applications such as people monitoring, surveillance, healthcare, HAR, and sleep monitoring \cite{Yildirim, Jun}. These approaches are less invasive and provide enhanced privacy and safety compared to traditional camera-based systems. Furthermore, multimodal frameworks that integrate information beyond CSI have been explored, enabling neural networks to leverage heterogeneous inputs such as video, RFID, and mmWave signals \cite{Islam}, thereby improving representation learning.

Many existing works rely exclusively on amplitude information, neglecting the rich content embedded in the phase component due to its sensitivity to hardware offsets, phase distortions, and environmental interference \cite{Wang}, \cite{LiTHAT}. However, disregarding phase information limits the full exploitation of CSI data, particularly in Delay–Doppler-based HAR, where motion dynamics provide highly discriminative features. A further challenge arises in multi-user and interaction-heavy settings: when several people act simultaneously, their reflections superimpose and one person may partially block the propagation paths between transmitter and receiver, degrading recognition. Several public datasets deliberately adopt such interaction-rich setups \cite{Huang}, where reliably identifying each activity is particularly difficult. Accordingly, a central objective of this work is to recognize activities robustly under signal superposition and interaction-heavy conditions while remaining effective across different environments.

Several studies have focused on developing robust preprocessing techniques within the field of Wi-Fi sensing to mitigate environmental noise and improve feature extraction. Wang et al. \cite{Wang} introduced the XRF55 dataset, a large-scale multimodal benchmark for activity recognition. We adopt XRF55 as our primary baseline, training our models exclusively on its Wi-Fi CSI component to evaluate performance in interaction-heavy scenarios. Multi-person interaction, where multiple subjects act simultaneously, remains a recurring challenge; Islam et al. \cite{Islam} addressed this using the Wi-MIR dataset and the CSI-IRNet model. We use Wi-MIR in our evaluation under the same framework to ensure a fair comparison.

Due to the high sensitivity of raw CSI phase to environmental factors, numerous studies have modeled motion in the Delay–Doppler domain. Hasanzadeh et al. \cite{Hasanzadeh} proposed MORIC, which decomposes CSI into delay components and classifies the resulting Delay–Doppler series. Similarly, Valaee et al. \cite{Valaee} introduced the Doppler Radiance Field (DoRF), reconstructing a 3D latent motion representation from 1D Doppler projections. Regarding larger gestures, Guo et al. \cite{Guo} proposed a SISO Self-Referencing Cross-Correlation (SRCC) method, featuring Doppler-ambiguity suppression and a MobileViT-XXS backbone.

Label scarcity has motivated self-supervised approaches: Xu et al. \cite{Dingchang} proposed a dual-stream contrastive framework that learns spatiotemporal features from raw CSI via mutual-information maximization. To counter environment-induced variability, Zhang et al. \cite{Ai} proposed CSI-GLSTN, which treats subchannels as graph nodes to learn a location-independent structure. For cross-domain gesture recognition, Liu et al. \cite{Liu} combined Doppler-based multi-angle images with a CBAM- and self-attention-augmented ResNet-18. Lastly, our prior work, IBIS \cite{Fernandes}, couples an Inception–BiLSTM feature extractor with an SVM post-processing stage tuned by grid search for cross-scenario evaluations.

A complementary line of research seeks to combine multiple CSI representations within a single model. In this direction, Li et al. \cite{LiTHAT} proposed THAT, a two-stream convolution-augmented Transformer for human activity recognition. Similarly, Lim et al. \cite{Lim} combined amplitude, phase-difference, and Doppler-shift representations for activity recognition. In a related manner, Delay–Doppler decompositions such as MORIC \cite{Hasanzadeh} explicitly exploit the channel's velocity structure but do not retain a parallel time-domain amplitude stream.

Table \ref{tab:contribution} summarizes the most recent studies related to Wi-Fi sensing discussed in this section, highlighting their integration of key components either directly or via alternative data modalities. It is worth emphasizing that most current approaches focus on specific preprocessing structures combined with neural network optimization strategies to maximize accuracy.

In this work, we address these challenges by proposing WiFuse, a unified pipeline that fuses two physically complementary CSI representations, a denoised time-domain amplitude stream and a Delay–Doppler stream derived from the sanitized phase, within a single hybrid spatio-temporal architecture. In contrast to prior approaches that rely on a single representation, or that fuse multiple views of the same amplitude tensor, WiFuse jointly learns from temporal signal energy and motion-induced velocity dynamics. To this end, a ResNet backbone is first trained and its learned weights are then used to initialize a hybrid ResNet–TCN network through transfer learning.

The main contributions of this work are summarized as follows:

\begin{itemize}
    \item We propose a dual-stream CSI representation that fuses a denoised time-domain amplitude stream with a Delay–Doppler stream derived from the sanitized phase, jointly exploiting temporal signal energy and motion-induced velocity dynamics;
    \item We design a hybrid ResNet–TCN architecture with channel and spatio-temporal attention, trained through a decoupled two-stage transfer learning strategy that first optimizes the spatial–spectral backbone and then fine-tunes the temporal modeling blocks;
    \item We conduct extensive experiments on two datasets to evaluate the effectiveness and generalization capability of the proposed pipeline, particularly in challenging multi-user interaction scenarios;
    \item We analyze and compare the computational cost of the proposed model against existing approaches reported in the literature.
\end{itemize}

The remainder of this paper is organized as follows. Section \ref{sec:proposta} presents the neural network modeling concepts, the Delay–Doppler formulation, and the mathematical foundations, as well as detailing the proposed WiFuse pipeline. 
Section \ref{sec:proposed_archtecture} explains in detail the proposed neural network architecture, its layers, and the algorithmic definitions. Section \ref{sec:avaliacao} reports the experimental results across different datasets, including confusion matrices and performance gain analyses in comparison with existing methods. Finally, Section \ref{sec:conclusao} concludes the paper and outlines directions for future research.

\begin{table*}[hbt!]
\small 
\caption{Summary of related works and the proposed approach.}
\label{tab:contribution}
\centering
\setlength{\tabcolsep}{3pt} 
\renewcommand{\arraystretch}{1.1} 
\begin{tabularx}{\textwidth}{p{2.5cm} c c p{2.8cm} p{2.5cm} X X}
\toprule
\textbf{Authors} & \textbf{Cl.} & \textbf{Year} & \textbf{Proposed} & \textbf{Model} & \textbf{Data Type} & \textbf{Contribution} \\
\midrule
Wang et al. \cite{Wang} & 55 & 2024 & Multimodal Dataset & ResNet 1D/2D & Amplitude & Deep Mutual Learning \\
Islam et al. \cite{Islam} & 17 & 2024 & MPI Dataset & CSI-IRNet & Amp. + Phase & Interaction Learning \\
Hasanzadeh \cite{Hasanzadeh} & 4 & 2025 & MORIC & MLP + Kernels & Delay--Doppler & Gen. Accuracy \\
Valaee et al. \cite{Valaee} & 4 & 2025 & DoRF & MLP + Kernels & Delay--Doppler & Radiance Field \\
Guo et al. \cite{Guo} & 9 & 2025 & WiDFS 3.0 & MobileViT-XXS & Delay--Doppler & Low-comp. SRCC \\
Xu et al. \cite{Dingchang} & 276 & 2023 & DualConFi & Custom MLP & Amp. + Phase & Contrastive Learning \\
Zhang et al. \cite{Ai} & 6 & 2024 & CSI-GLSTN & CSI-GLSTN & Amp. + Phase & Temporal Learning \\
Liu et al. \cite{Liu} & 6 & 2025 & Cross-Domain & ResNet-18 & Doppler & Enhanced Accuracy \\
Fernandes \cite{Fernandes} & 9 & 2025 & IBIS & Inception-BiLSTM & Doppler & Generalization \\
Li et al. \cite{LiTHAT} & 7 & 2021 & THAT & Transformer & Amplitude & Conv-Augmented \\
Quy et al. \cite{Quy} & -- & 2025 & Ph-Amp-Att & CNN + Att. & Amp. + Phase & Attention Fusion \\
Lim et al. \cite{Lim} & -- & 2023 & Score-Fusion & Multi-Stream CNN & Amp.+Phase+Doppler & Score-level Fusion \\
\textbf{WiFuse*} & 55/17 & 2026 & \textbf{WiFuse} & \textbf{ResNet--TCN} & \textbf{Amp. + Delay-Doppler} & \textbf{Hybrid Fusion} \\
\bottomrule
\end{tabularx}
\begin{flushleft}
\footnotesize{Note: The symbol * represents our proposal WiFuse.}
\end{flushleft}
\end{table*}

\section{Background}
\label{sec:proposta}

This section presents the main concepts related to Wi-Fi sensing, as well as the mathematical formulations involving amplitude and Delay-Doppler representations. It also describes the neural network architectures adopted in our solution.

\subsection{Channel State Information}

For human activity recognition, Channel State Information (CSI) plays a crucial role in detecting movements and actions, as it captures fine-grained variations in the wireless channel caused by human motion.

In wireless communication systems, the Channel Impulse Response (CIR) is commonly used to describe and characterize the wireless channel, and is expressed as~\cite{YangRSSI}:

\begin{equation}
    h(t) = \sum_{l=1}^{N} \alpha_l e^{-j \beta_l} \delta(t - \tau_l)
\end{equation}

where $N$ denotes the total number of propagation paths, $\delta(\cdot)$ is the Dirac delta function, and $\alpha_l$, $\beta_l$, and $\tau_l$ denote, respectively, the amplitude attenuation, phase offset, and propagation delay of the $l$-th path.

In the frequency domain, multipath propagation produces frequency-selective fading, characterized by the Channel Frequency Response (CFR), which comprises an amplitude and a phase frequency response. The CFR is the Fourier transform of the CIR with respect to the delay variable, so that the two constitute a Fourier transform pair; for a band-limited OFDM system, the CSI provides sampled observations of the CFR at the subcarrier frequencies.

CSI reflects the fundamental characteristics of the physical-layer wireless channel during signal propagation, typically measured on individual Orthogonal Frequency-Division Multiplexed (OFDM) subcarriers. It provides detailed information about channel conditions and performance. A set of CSI values can be obtained from received data packets, each capturing the amplitude and phase of an OFDM subcarrier and evolving across packets. For subcarrier $k$ at packet index $n$, the complex CSI value is given by:

\begin{equation}
    H(k,n) = \| H(k,n) \| \, e^{j \angle H(k,n)}
\end{equation}

where $\| H(k,n) \|$ and $\angle H(k,n)$ denote, respectively, the amplitude and phase of subcarrier $k$ at packet index $n$.

The CSI is estimated per subcarrier from known preamble symbols (e.g., by least squares) and, in the multi-antenna setting, is collected over all transmit--receive antenna pairs; stacking these measurements across subcarriers and packets yields the CSI tensor processed in this work.

\subsection{Preprocessing Dual-Stream Data}

CSI data preprocessing is of fundamental importance for effective analysis and utilization in deep learning–based human activity recognition tasks.

Initially, raw CSI data are collected, and a phase sanitization process \cite{Min, Diaz} is applied, since raw phase information cannot be directly exploited due to hardware-induced offsets and random phase distortions. To address these issues, the following procedures are performed:

\begin{itemize}
    \item \textbf{Phase Unwrapping}: Prevents the phase from exceeding the $[-\pi, \pi]$ interval, which would otherwise introduce discontinuities and render the signal unusable \cite{Dang};
    \item \textbf{Index Centralization}: Centers the subcarrier indices around zero to facilitate slope estimation;
    \item \textbf{Linear Regression}: Estimates the slope and intercept of the phase variation across subcarriers using the least-squares method;
    \item \textbf{Phase Reconstruction}: Mathematically reconstructs a clean phase signal, removing offsets, discontinuities, and interference-induced noise.
\end{itemize}

The time-domain amplitude stream is obtained directly from the raw CSI magnitude, independently of the phase sanitization, by computing the complex magnitude of each subcarrier for every packet:

\begin{equation}
    \| H(k,n) \| = \sqrt{\Re\{H(k,n)\}^2 + \Im\{H(k,n)\}^2}
\end{equation}

To mitigate noise, interference, and oscillations unrelated to human motion, temporal smoothing is applied with a moving-average filter of 30 packets along the packet axis. The window length is chosen relative to the CSI sampling rate so as to suppress rapid channel fluctuations while preserving the lower-frequency macroscopic envelopes of human activity; the corresponding time span for each dataset is specified in Section~\ref{sec:avaliacao}.

The resulting four-dimensional amplitude stream (sample $\times$ antenna $\times$ subcarrier $\times$ packet) is then reshaped by flattening the antenna and subcarrier dimensions into a single feature vector per packet, enabling direct input to the neural network.

For the Delay–Doppler stream, the sanitized complex CFR $\tilde{H}(k,n) = \| H(k,n) \| \, e^{\,j \hat{\theta}(k,n)}$ is used, where $\hat{\theta}(k,n)$ is the reconstructed (sanitized) phase. An Inverse Fast Fourier Transform (IFFT) is first applied across the subcarrier dimension, mapping the signal into the delay domain and separating the multiple propagation paths; since propagation distance is directly related to delay, this isolates the reflections associated with human motion. A Fast Fourier Transform (FFT) is then applied along the packet (time) axis, converting the temporal phase variations into Doppler frequencies. The joint Delay–Doppler representation is therefore obtained as

\begin{equation}
    D(\tau, f_d) = \mathcal{F}_{n}\big\{ \mathcal{F}^{-1}_{k}\{ \tilde{H}(k,n) \} \big\},
    \label{eq:delay_doppler}
\end{equation}

where $\tau$ and $f_d$ denote the delay and Doppler bins, and $\mathcal{F}_{n}$ and $\mathcal{F}^{-1}_{k}$ are the FFT along the packet axis and the IFFT across subcarriers, respectively. The spectrum is shifted so that the zero-Doppler component is centered, and its magnitude is compressed with a logarithmic scale to reduce the wide dynamic range of the reflected power.

The z-score normalization is applied on a per-sample basis to ensure that the data are centered and properly scaled. This step is essential for stable optimization and efficient training of deep neural networks.

At this stage, the amplitude and Delay–Doppler representations are concatenated, introducing the dual-stream concept. This strategy enables the neural network to learn complementary patterns from two distinct CSI modalities simultaneously, which would not be possible when using a single representation. The two streams are defined as follows:
\begin{itemize}
    \item \textbf{Amplitude / Time Stream}: Operates on the denoised raw-magnitude amplitude, focusing on energy variations over time and capturing temporal patterns related to human motion dynamics and physical obstacles;
    \item \textbf{Delay–Doppler / Frequency Stream}: Operates on the sanitized phase information by applying a two-dimensional Fourier Transform (2D-FFT), emphasizing spectral signatures, multipath separation, and motion-related Doppler patterns.
\end{itemize}

The combination of these two modalities constitutes a CSI feature-fusion strategy, producing a three-dimensional input tensor of shape $(2 \times F \times T)$. The first dimension indexes the two CSI modalities (channel~0, the denoised amplitude; channel~1, the Delay–Doppler representation derived from the sanitized phase); the second dimension, of size $F = N_{\ell}\,N_{\mathrm{sc}}$, stacks the spatial--frequency features formed by the $N_{\ell}$ transmit--receive antenna links across $N_{\mathrm{sc}}$ subcarriers; and the third dimension, of length $T$, indexes the packet (slow-time) axis and captures the temporal evolution of the channel. For input to the network, the two modality channels are concatenated along the feature axis, yielding $2F$ input channels over the $T$-packet axis, so that all subsequent layers operate on a joint cross-domain feature space. The dataset-specific values of $N_{\ell}$, $N_{\mathrm{sc}}$, and $T$ are reported in Section~\ref{sec:avaliacao}. This dual-stream representation enhances the model's ability to exploit both the temporal and spectral characteristics inherent in CSI data, improving robustness and recognition performance.

\subsection{Delay–Doppler Representation}

In OTFS-based and Delay–Doppler signal processing, a time-varying multipath channel is represented on a two-dimensional Delay–Doppler grid, on which moving scatterers appear as compact, physically interpretable peaks rather than as the smeared responses seen in the time--frequency domain \cite{Xia}. WiFuse adopts this view: the sanitized CSI, naturally indexed by subcarrier (frequency) and packet (slow time), is mapped to the Delay–Doppler domain by the two-dimensional transform of Eq.~(\ref{eq:delay_doppler})---an inverse FFT across the $M$ subcarriers (frequency $\rightarrow$ delay) followed by an FFT across the $N$ packets (slow time $\rightarrow$ Doppler), analogous to the (inverse) symplectic finite Fourier transform that defines the OTFS Delay–Doppler representation~\cite{Hadani}.

The resulting grid has $M$ delay bins and $N = 1000$ Doppler bins. While the physical boundaries are established by the hardware during acquisition, the final grid dimensions are standardized via digital resampling to ensure a uniform tensor shape. The native delay resolution and unambiguous delay span are governed by the channel bandwidth:
\begin{equation}
\Delta\tau = \frac{1}{B}, \qquad \tau_{\max} = \frac{1}{\Delta f},
\label{eq:delay_res}
\end{equation}
where $B$ is the occupied bandwidth and $\Delta f$ the subcarrier spacing. Meanwhile, the unambiguous Doppler span depends strictly on the physical CSI sampling period $T_s$. Following the Fourier transforms, the Doppler spectrum is resampled to a fixed target of $N$ bins, yielding a constant digital Doppler resolution and a physical span defined by:
\begin{equation}
\Delta\nu = \frac{1}{T_f} = \frac{1}{N T_s}, \qquad |\nu|_{\max} = \frac{1}{2 T_s},
\label{eq:doppler_res}
\end{equation}
where $T_f = N T_s$ represents the standardized effective observation duration of the resampled grid. A Doppler shift $\nu$ maps to a radial velocity through $v = \nu c / (2 f_c)$, resulting in a fixed velocity resolution of $\Delta v = c/(2 f_c N T_s)$.

These quantities are perfectly optimized for the dataset processing pipeline. Regardless of the raw recording durations, the physical sampling rate of $200$~Hz ($T_s = 5$~ms) establishes a fixed unambiguous Doppler span of $\pm 100$~Hz. By digitally resampling the Doppler domain to $N = 1000$ bins, Eq.~(\ref{eq:doppler_res}) enforces a uniform resolution of $\Delta\nu = 0.2$~Hz (equivalent to a standardized $T_f = 5$~s). With a carrier $f_c = 5.64$~GHz, this configuration yields a constant velocity resolution of $\Delta v \approx 5.3$~mm/s over an unambiguous range of about $\pm 2.7$~m/s. This finely resolves the radial velocities of human activities---from subtle limb motion (a few cm/s) up to the brisk whole-body movements present in the datasets (walking and running). On the delay axis, the $20$~MHz Wi-Fi channel of the Intel 5300 front-end provides a native $\Delta\tau = 50$~ns ($\approx 15$~m path-length resolution), while $\tau_{\max}$ captures all relevant multipath components unambiguously. Concurrently, the time-domain amplitude stream is also resampled to $N = 1000$ points, effectively normalizing the temporal scale of each activity sequence. This dual-channel resampling mechanism ensures a symmetric, fixed-size input tensor that preserves the essential motion information exploited by the network.


\subsection{Spatial and Temporal Building Blocks}
The Residual Network (ResNet) \cite{He} eases the optimization of deep networks through residual blocks, in which a shortcut connection adds the block input to its output so that the network learns residual mappings, improving gradient flow and mitigating vanishing gradients. WiFuse uses ResNet as a one-dimensional spatial--spectral extractor over the 540 fused channels, with an integrated channel-attention mechanism (Section~\ref{sec:proposed_archtecture}) that emphasizes the most discriminative channels; its 1D residual configuration is detailed in Table~\ref{tab:resnet_tcn_arch}.

Temporal Convolutional Networks (TCNs) \cite{Bai} model sequences with dilated causal convolutions and residual connections \cite{Reiter}, capturing long-range dependencies more efficiently and stably than recurrent networks, since their convolutions are fully parallelizable and free of the vanishing-gradient issues of RNNs. Causality makes the output at step $n$ depend only on current and past inputs, $y_n = f(x_1,\dots,x_n)$ for $n=1,\dots,T$ (with $T$ the sequence length), while dilation enlarges the receptive field without extra layers: for an input $\mathbf{x}\in\mathbb{R}^T$ and a length-$K$ filter, the dilated convolution at position $p$ is $F(p)=\sum_{i=0}^{K-1} f(i)\,x_{p-d\cdot i}$, with $d$ the dilation factor. Stacking blocks with exponentially increasing $d$ expands the receptive field geometrically, letting the TCN model long CSI sequences compactly.

\section{Proposed WiFuse Framework}
\label{sec:proposed_archtecture}

\begin{figure*}[ht]
    \begin{center}
      \includegraphics[width=1.0\textwidth]{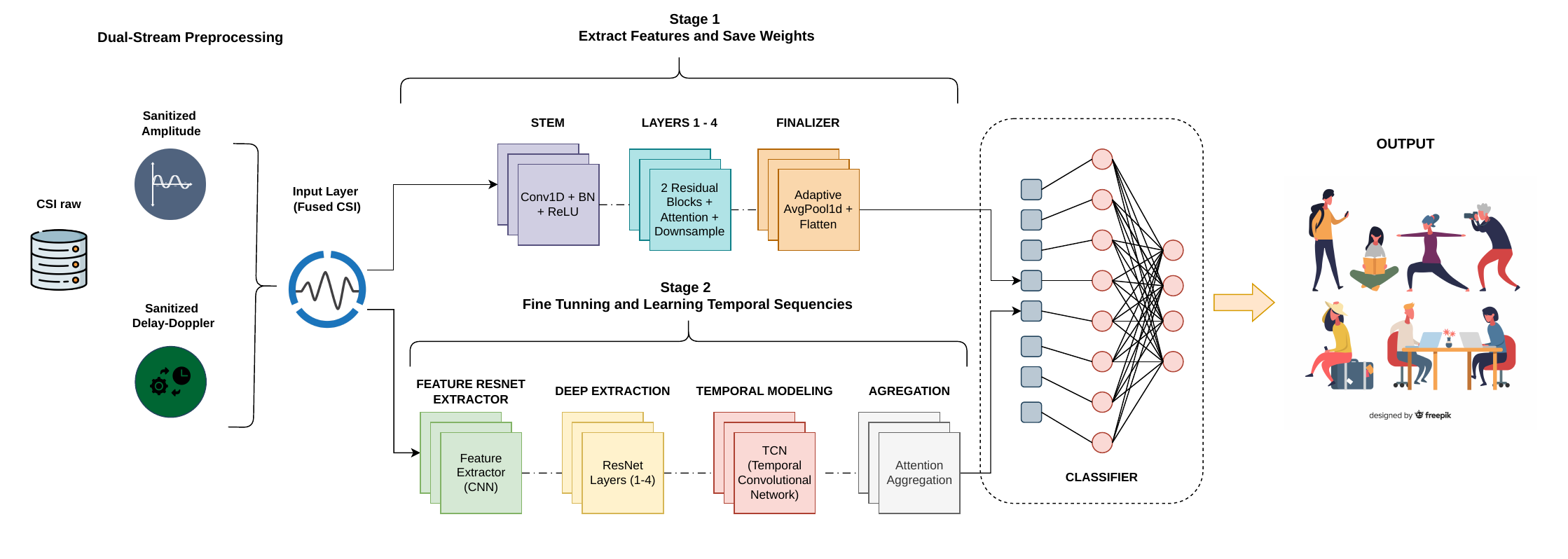}
        \caption{Illustration of the WiFuse framework showing the preprocessing stages used for feature extraction, as well as the application of transfer learning from the ResNet network to the hybrid ResNet–TCN neural architecture.}
        \label{fig:resnet_tcn}
    \end{center}    
\end{figure*}

The proposed WiFuse pipeline is illustrated in Fig.~\ref{fig:resnet_tcn}. Its defining feature is the \emph{feature-level} fusion of two physically complementary streams, the denoised time-domain amplitude and the sanitized-phase Delay–Doppler representation, which are concatenated into a single dual-domain input and learned jointly, rather than processed in isolation or combined only at the decision level. Spatial attenuation patterns and signal peaks are captured primarily from the amplitude stream, whereas motion-induced velocity dynamics and action-related delays are resolved from the Delay–Doppler stream. Two design elements tailor the network to this fused input: a ResNet backbone with channel attention for spatial--spectral calibration, followed by a TCN with spatio-temporal attention for long-range motion modeling, optimized through a decoupled two-stage transfer learning strategy.

The ResNet of Stage~1 ingests the fused dual-stream tensor, i.e., $2F$ channels ($F$ amplitude and $F$ Delay–Doppler features) over a temporal horizon of $T$ packets. The stem block captures initial signal variations, followed by Layers~1 to~4, which are residual blocks equipped with batch normalization and channel attention. These mechanisms assign higher weights to the feature channels containing useful motion-related information while suppressing noise and interference. Another important characteristic of these layers is their ability to detect rapid signal changes, which is crucial for identifying complex patterns commonly observed in human activities.

Finally, a global pooling layer is applied to extract compact representations, followed by a classifier with a softmax activation function to produce probabilistic outputs. Table~\ref{tab:resnet_tcn_arch} details all layers, parameters, and output shapes of the complete architecture.

In Stage 2, using the previously learned weights, a new neural network architecture, referred to as ResNet-TCN is constructed. In the first step, the input signal is processed by the ResNet backbone to extract spatial features learned in the previous stage. Layers 1 to 4 correspond to the ResNet blocks with channel attention and serve as the backbone for extracting spatial characteristics from the 540 input channels, as well as for feature calibration. The transferred channel attention modules recalibrate the feature channels, while a separate spatio-temporal attention block at the TCN output generates adaptive temporal weights, enabling the model to focus on the most relevant signal segments.

In the TCN module itself, progressively increasing dilations are employed, allowing the model to capture long-range temporal dependencies in motion signals and aggregate useful information for the activity recognition process. Table \ref{tab:resnet_tcn_arch} summarizes all network components and output shapes related to the implemented neural architecture.

\begin{table}[ht!]
\caption{ResNet-TCN Fusion Architecture with Temporal Attention}
\label{tab:resnet_tcn_arch}
\begin{center}
\begin{tabular}{p{4.2cm}|p{1.8cm}|p{1.4cm}}\toprule
\textbf{Block / Layer} & \textbf{Output Shape} & \textbf{Parameters} \\ \midrule
Input Layer (Fused CSI) & (540, 1000) & - \\ \midrule

\textbf{Feature Extractor (CNN)} & & \\
Conv1D(64$\times$7$\times$540, stride=2) & (64, 500) & 241,920 \\
BatchNormalization & (64, 500) & 128 \\
Activation (ReLU) & (64, 500) & - \\ \midrule

\textbf{ResNet Layers (1-4)} & & \\
Layer 1: 2$\times$ ResBlock(64) & (64, 500) & $\approx$ 50,688 \\
Layer 2: 2$\times$ ResBlock(128) & (128, 250) & $\approx$ 188,000 \\
Layer 3: 2$\times$ ResBlock(256) & (256, 125) & $\approx$ 745,000 \\
Layer 4: 2$\times$ ResBlock(512) & (512, 63) & $\approx$ 2,960,000 \\ \midrule

\textbf{Temporal Modeling (TCN)} & & \\
Block 1: Conv1D(512$\times$3$\times$512, d=1) & (512, 63) & 786,432 \\
Block 1: BatchNorm + ReLU & (512, 63) & 1,024 \\
Block 2: Conv1D(512$\times$3$\times$512, d=2) & (512, 63) & 786,432 \\
Block 2: BatchNorm + ReLU & (512, 63) & 1,024 \\
Block 3: Conv1D(512$\times$3$\times$512, d=4) & (512, 63) & 786,432 \\
Block 3: BatchNorm + ReLU & (512, 63) & 1,024 \\
Block 4: Conv1D(512$\times$3$\times$512, d=8) & (512, 63) & 786,432 \\
Block 4: BatchNorm + ReLU & (512, 63) & 1,024 \\ \midrule

\textbf{Attention Aggregation} & & \\
Conv1D(1$\times$1$\times$512) & (1, 63) & 513 \\
Softmax (Temporal) & (1, 63) & - \\
Weighted Sum (Dot Product) & (512,) & - \\ \midrule

\textbf{Classifier} & & \\
Dropout & (512,) & - \\
Linear (Fully Connected) & (55,) & 28,215 \\
\bottomrule
\end{tabular}
\end{center}
\raggedright
\footnotesize{Note: Output shape format is $(Channels, Time)$ or $(Vector Size,)$. d -  refers to dilation rate. The input time dimension ($T=1000$) reduces through the ResNet strides but is preserved in the TCN blocks due to padding. The final Attention module collapses the temporal dimension into a feature vector}
\end{table}

\subsection{Channel Attention}

To enable the application of the dual-stream concept and allow the network to extract relevant information, a channel attention module~\cite{CBAM} was implemented. Its primary function is to filter which components of the Wi-Fi signal are informative and which are predominantly noise. The structure of the channel attention mechanism is described as follows.

\begin{itemize}
    \item \textbf{Average Pooling}: Extracts statistical information from each signal channel, capturing the overall trend and general motion-related characteristics;
    \item \textbf{Max Pooling}: Identifies the strongest peaks of signal variation, which predominantly correspond to the displacement and movement of the subject;
    \item \textbf{Shared MLP}: Within the channel attention module, a lightweight shared multilayer perceptron (MLP) is employed to learn the correlations among channels and to emphasize their most salient characteristics.
\end{itemize}

This attention mechanism was incorporated to mitigate the effects of constant interference, reflective surfaces, multipath propagation, and hardware-induced offsets. By extracting complementary features from the dual-stream representation, the channel attention module enables the ResNet to focus on the most informative and discriminative signal components, thereby improving the robustness and effectiveness of the feature extraction process.

\subsection{Spatio-Temporal Attention}

In the proposed WiFuse framework, a Spatio-Temporal Attention (STA) mechanism is incorporated at the output of the final TCN layer. Although the TCN captures both short- and long-term temporal dependencies, recognition can be further improved by explicitly emphasizing the most informative time steps. Attention mechanisms provide precisely this capability: by learning a data-dependent weighting over the sequence, they highlight the relevant elements while suppressing the irrelevant ones~\cite{Vaswani}.

This mechanism is particularly important in Wi-Fi CSI-based human activity recognition, since human movement is not static but rather a temporal process characterized by sequential variations in subcarriers occurring in a specific order. The attention layer allows the model to identify and emphasize the most informative moments within a CSI sequence, which are often critical for distinguishing between similar activities.

The primary function of this layer is to capture long-term temporal dependencies through dilated temporal convolutions, while preserving the inherent sequential structure of the signal. While the ResNet backbone employs a channel attention mechanism to enhance discriminative spatial–spectral features, the TCN processes the CSI data as a temporal series, fully exploiting its rich temporal dynamics.

This architectural design benefits the network by enabling it not only to model temporal dependencies but also to identify the exact moments at which specific movements occur within each sample, which is essential for robust recognition of complex and joint human activities.

\subsection{Training Procedure}

WiFuse is trained in two stages, both using the AdamW optimizer under mixed precision (FP16) for 100 epochs, after the dual-stream features are concatenated and z-score normalized. The data are split 80\%/20\% into training and test sets, with a held-out validation subset (10\% of the training partition) used for learning-rate scheduling and best-model selection, while the test partition is reserved for final evaluation.

In Stage~1, the ResNet backbone with channel attention is trained from scratch with a learning rate of $0.001$ and a batch size of $32$; the rate is halved whenever the validation accuracy does not improve for five consecutive epochs (ReduceLROnPlateau), and the best weights are retained. In Stage~2, these weights initialize the hybrid ResNet--TCN, which is fine-tuned with layer-specific learning rates---$2\times10^{-6}$ for the pre-trained CNN layers, $2\times10^{-5}$ for the pre-trained channel-attention layers, and $2\times10^{-4}$ for the newly initialized TCN, spatio-temporal attention, and classifier---together with a weight decay of $0.02$, a batch size of $64$, label smoothing of $0.1$, and a cosine-annealing schedule with warm restarts whose periodic restarts help the optimizer escape poor local minima. The complete set of hyperparameters is summarized in Table~\ref{tab:params_comparison}.

\section{Experimental Evaluation}
\label{sec:avaliacao}

This section characterizes the experimental evaluation of the proposed WiFuse framework on both datasets and provides comparative benchmarks against other neural networks and processing pipelines. Furthermore, it presents experiments involving ablation studies, domain adaptation, and cross-domain evaluation to assess the effectiveness of the proposed framework. To avoid any potential bias in the results, a total of ten simulations were performed for all experiments involving the XRF55 and Wi-MiR datasets, and the average values were calculated across all simulations.

\subsection{XRF55 Dataset}

The XRF55 dataset was proposed by Wang et al. \cite{Wang} as a multimodal dataset comprising the following sensing modalities: Wi-Fi CSI, RFID, and millimeter-wave (mmWave) radar.

XRF55 comprises 23 RFID tags (922.38~MHz), nine Wi-Fi links (5.64~GHz), one mmWave radar (60--64~GHz), and one Azure Kinect (RGB/depth/IR), recording 55 action classes from 39 participants for 42{,}900 valid samples ($>$59 hours). The Wi-Fi CSI---the only modality used in this work---was captured with an Intel 5300 NIC (a transmitter with three antennas and three receivers placed at the corners of a rectangular area) over 30 OFDM subcarriers. In the notation of Section~\ref{sec:proposta}, the nine transmit--receive links (three transmit $\times$ three receive antennas) over $N_{\mathrm{sc}}=30$ subcarriers give $F = N_{\ell}\,N_{\mathrm{sc}} = 9 \times 30 = 270$ spatial--frequency features (equivalently, 90 per receive antenna, the per-antenna count referenced earlier). The CSI is sampled at $200$~Hz, so each $5$~s recording yields $T = 1000$ packets, and the 30-packet smoothing window therefore spans $150$~ms. The 55 classes are grouped into five categories---Human--Object Interaction, Human--Human Interaction, Fitness, Body Motion, and Human--Computer Interaction.

Four physical scenarios are provided: \emph{Scene~1}, a furnished living room with 30 participants performing all 55 activities; and \emph{Scenes~2--4}, each with three participants---an entrance hall with reflective walls (Scene~2), an empty room containing only the experimental equipment (Scene~3), and an empty room with windows and reflective surfaces (Scene~4). Raw CSI was filtered with the Linux 802.11n CSI Tool~\cite{Halperin}, and only the Wi-Fi CSI modality is used in our experiments.

\subsection{Wi-MIR Dataset}

The second dataset, Wi-MIR \cite{Islam}, targets multi-person interaction. It was recorded in an 8~m $\times$ 6.1~m furnished indoor room with a NETGEAR Nighthawk R7000 access point and an Intel 5300 NIC receiver (three antennas) in a line-of-sight configuration 7.4~m apart, using the Linux 802.11n CSI Tool. It comprises 3{,}740 trials (20 per interaction) from 11 subject pairs over 17 joint-activity classes, with CSI of dimensions $I \times N_{\mathrm{rx}} \times N_{\mathrm{tx}} \times N_{\mathrm{sc}}$ ($N_{\mathrm{sc}}=30$, $N_{\mathrm{tx}}=N_{\mathrm{rx}}=3$, and $I$ packets per trial). With $N_{\mathrm{tx}}=N_{\mathrm{rx}}=3$ and $N_{\mathrm{sc}}=30$, Wi-MIR has the same nine transmit--receive links and therefore the same $F=270$ spatial--frequency features as XRF55 (90 per receive antenna), which is what allows the identical network to be applied; each trial is brought to the network's fixed temporal length of $T=1000$ packets.

It is important to emphasize that the primary focus of the Wi-MIR dataset is on joint actions between participants, which leads to a significantly higher level of difficulty for activity recognition and poses greater challenges for neural network–based classification models.

\subsection{XRF55 Dataset Validation}

\begin{figure*}[ht!]
    \centering
    
    \begin{minipage}[b]{0.24\textwidth}
        \centering
        \includegraphics[width=\textwidth]{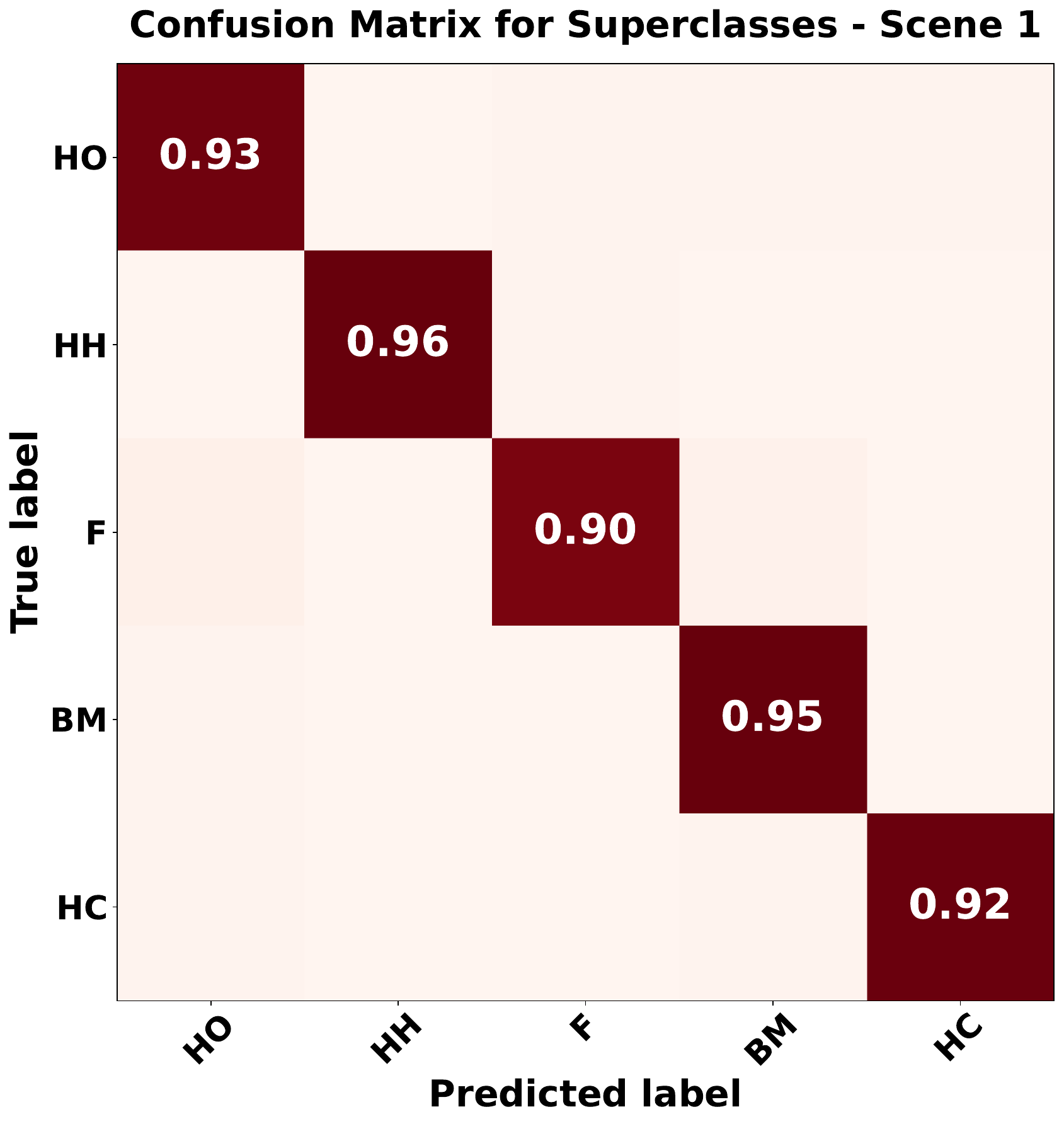}
        \caption*{\small (a) Scene 1 - 93.24\%}
        \label{fig:a}
    \end{minipage}
    \hfill
    \begin{minipage}[b]{0.24\textwidth}
        \centering
        \includegraphics[width=\textwidth]{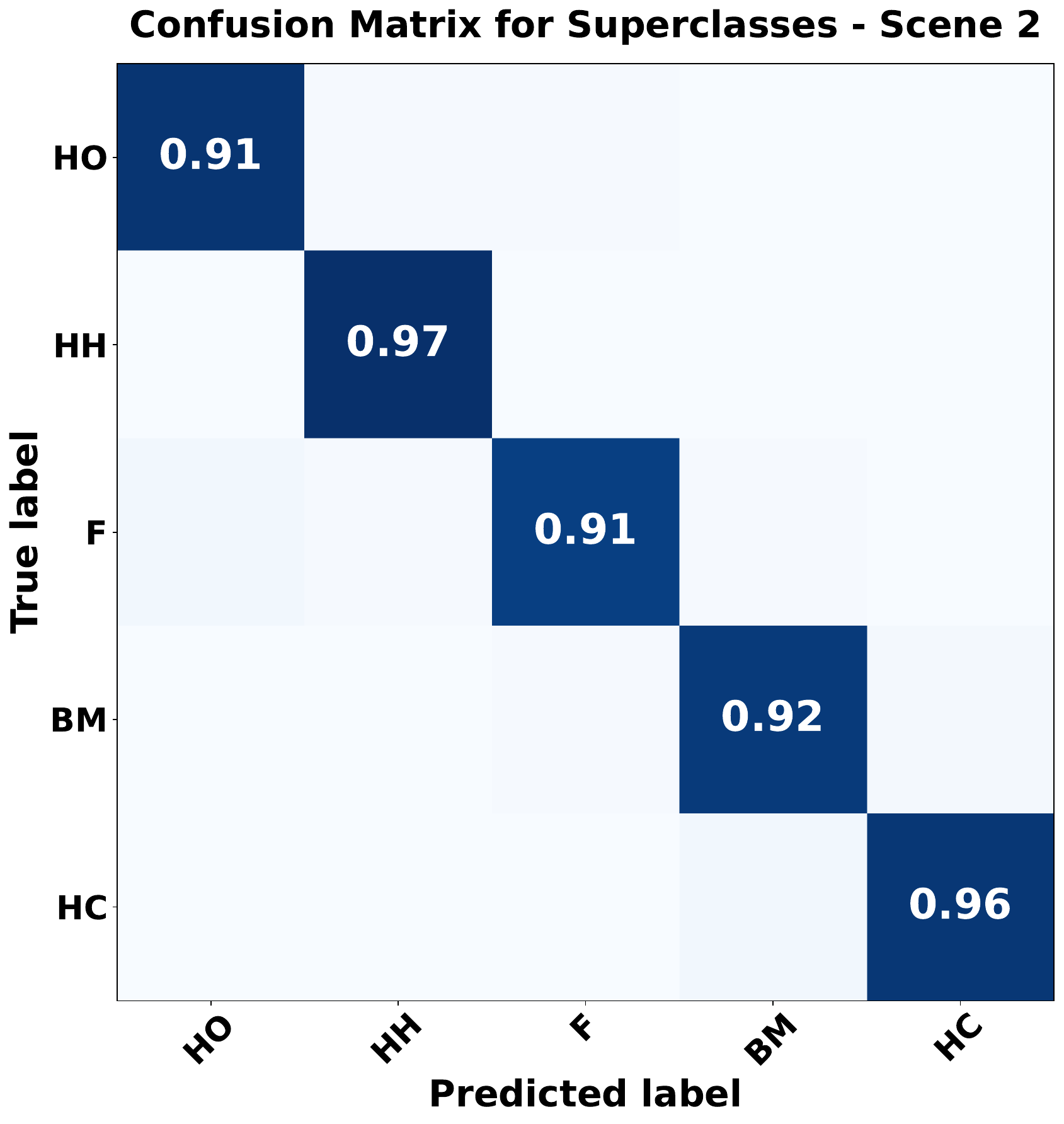}
        \caption*{\small (b) Scene 2 - 93.45\%}
        \label{fig:b}
    \end{minipage}
    \hfill
    \begin{minipage}[b]{0.24\textwidth}
        \centering
        \includegraphics[width=\textwidth]{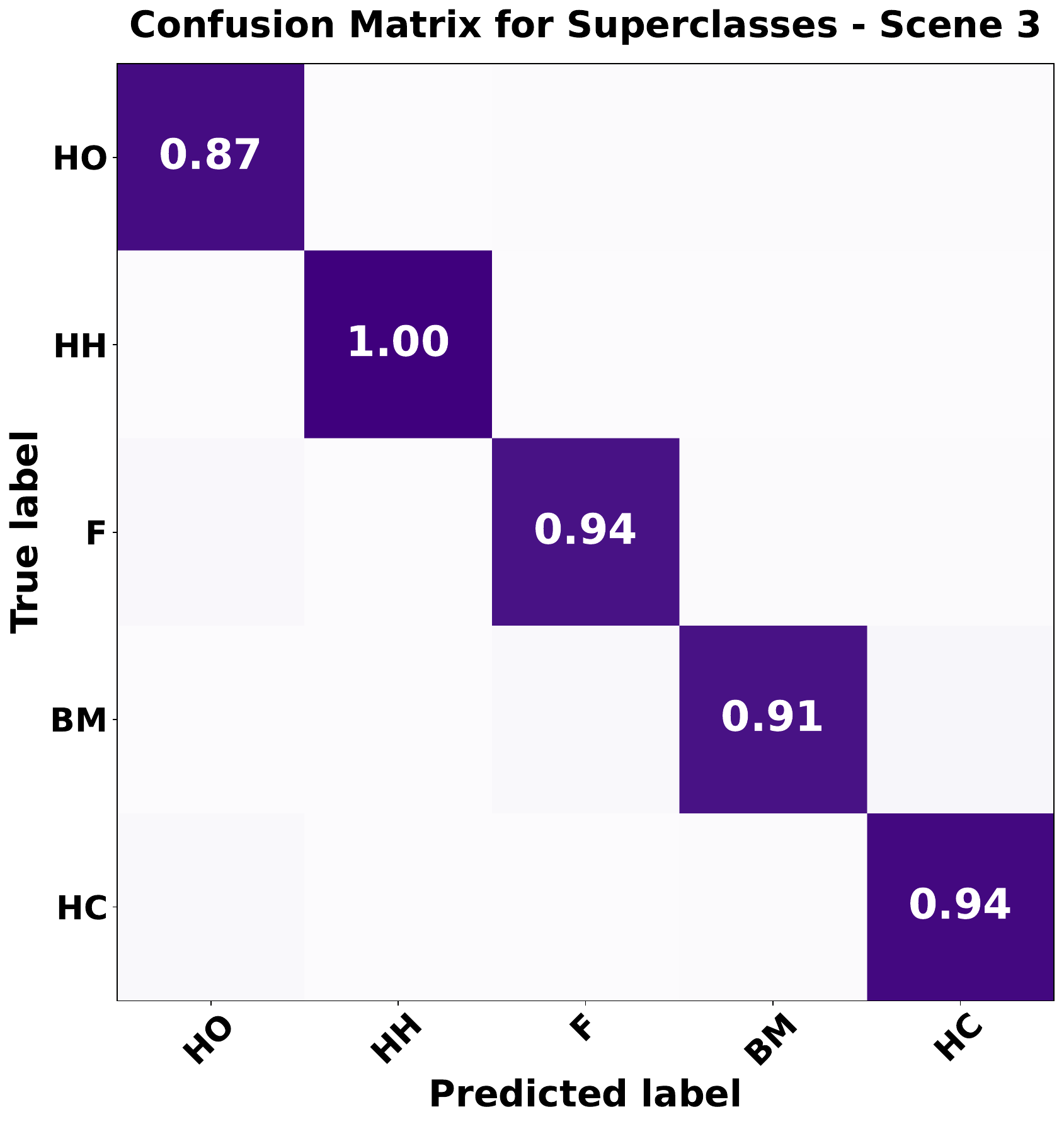}
        \caption*{\small (c) Scene 3 - 93.27\%}
        \label{fig:c}
    \end{minipage}
    \hfill
    \begin{minipage}[b]{0.24\textwidth}
        \centering
        \includegraphics[width=\textwidth]{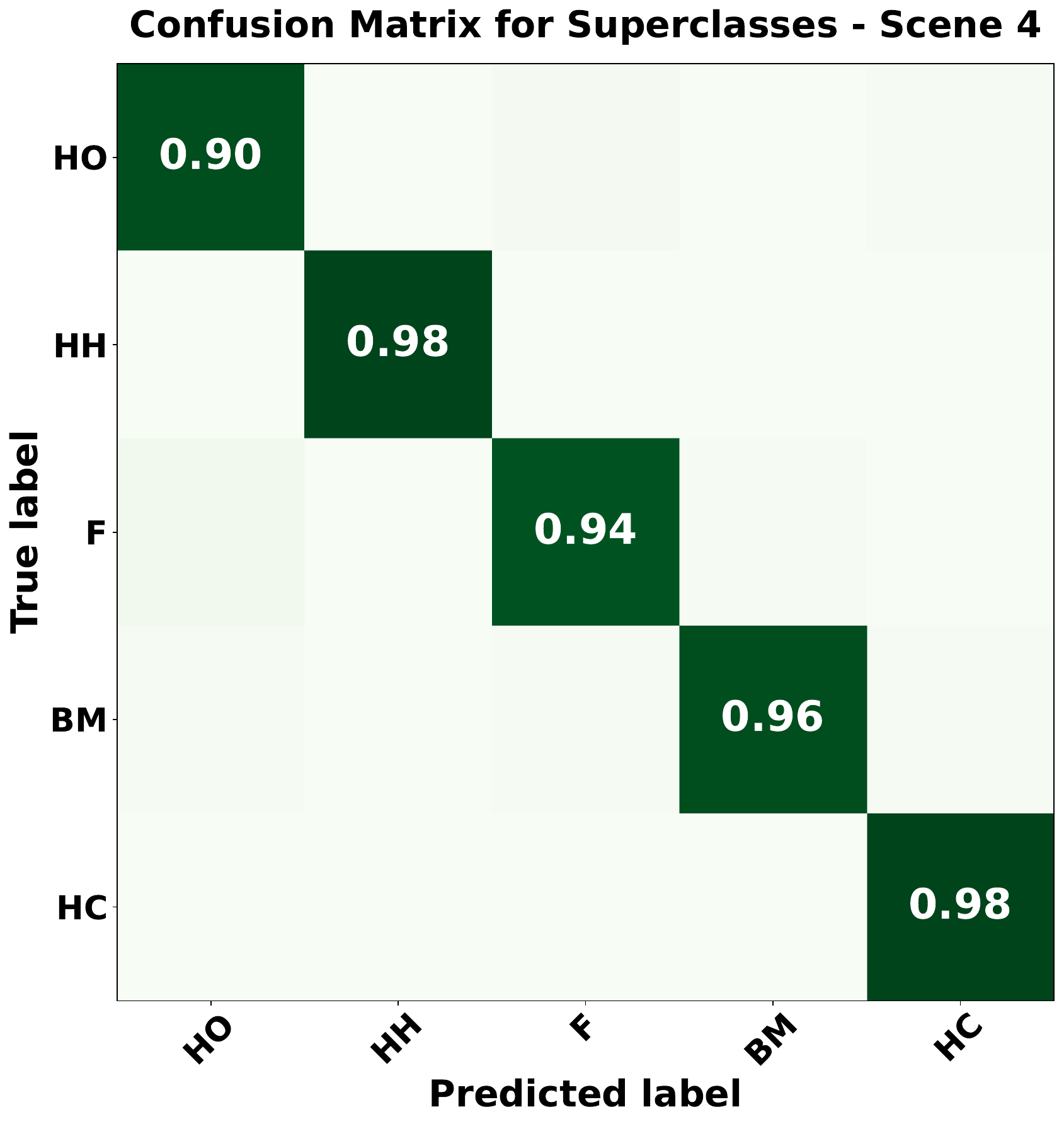}
        \caption*{\small (d) Scene 4 - 95.28\%}
        \label{fig:d}
    \end{minipage}

    \vspace{0.2cm}

    \begin{flushleft}
        \footnotesize \textit{Note: HO - Human-Obj, HH - Human-Hum, F - Fitness, BM - Body Motion, and HC - Hum-Comp}.
    \end{flushleft}
    
    \caption{Performance evaluation across multiple environments. The confusion matrices validate the model's robustness in (a) Scene 1, (b) Scene 2, (c) Scene 3, and (d) Scene 4, validating the model's robustness across all activity superclasses}
    \label{fig:conf_mx}
    
\end{figure*}

To validate the data from the XRF55 dataset, several experiments were conducted considering the four distinct scenarios defined in the dataset. Fig.~\ref{fig:conf_mx} presents the obtained accuracy results. It is worth noting that the XRF55 dataset comprises 55 distinct classes; therefore, presenting all classes individually would compromise result visualization. To address this issue, the concept of superclasses was adopted. Specifically, the activities were evaluated according to the five predefined groups: Human–Object Interaction, Human–Human Interaction, Fitness, Body Motion, and Human–Computer Interaction.

Another point that should be mentioned concerns the network structure used. The accuracy results presented were obtained from the fine-tuning process through transfer learning of the hybrid ResNet–TCN network. As for the script dedicated to the ResNet architecture and its training procedure, it will be evaluated later in the ablation study section.

Table~\ref{tab:params_comparison} presents the parameters used in this experiment.

\begin{table}[hbt!]
\caption{Hyperparameters and Experimental Settings for ResNet and ResNet-TCN in the WiFuse framework}
\label{tab:params_comparison}
\centering
\resizebox{\columnwidth}{!}{%
\begin{tabular}{lll}
\toprule
\textbf{Parameter} & \textbf{ResNet} & \textbf{ResNet-TCN} \\
\midrule
Architecture & ResNet & ResNet-TCN \\
Input Channels & 540 (Dual Stream) & 540 (Dual Stream) \\
Optimizer & AdamW & AdamW \\
Learning Rate & $1 \times 10^{-3}$ & $2 \times 10^{-6}$ to $2 \times 10^{-4}$ (Split LR) \\
LR Scheduler & ReduceLROnPlateau & CosineAnnealingWarmRestarts \\
Weight Decay & $1 \times 10^{-2}$ & $2 \times 10^{-2}$ \\
Number of Epochs & 100 & 100 \\
Batch Size & 32 & 64 \\
Loss Function & CrossEntropyLoss & CrossEntropyLoss (Label Smoothing 0.1) \\
Precision & Mixed Precision (FP16) & Mixed Precision (FP16) \\
Data Augmentation & Gaussian Noise (0.005) & Gaussian Noise (0.008) + Time Masking \\
Activation Function & ReLU & ReLU \\
\bottomrule
\end{tabular}}
\begin{flushleft}
\footnotesize{Note: Comparative training setup, highlighting the regularization techniques and learning-rate schedules used for each stage.}
\end{flushleft}
\end{table}

In Scene~1 (Fig.~\ref{fig:conf_mx}a), Body Motion attains the highest superclass accuracy (95\%) and Human--Human Interaction reaches 96\%, owing to the rich Doppler signatures of macro-activities such as jumping and running; the remaining superclasses exceed 90\%, with Fitness lowest (e.g., Boxing, Tai Chi) due to its constrained, short-duration movements. In the three-subject Scene~2 (Fig.~\ref{fig:conf_mx}b), despite the reflective entrance hall, Human--Human Interaction peaks at 97\% while Human--Object Interaction and Fitness are lowest (91\%). In the empty Scene~3 (Fig.~\ref{fig:conf_mx}c), Human--Human Interaction reaches 100\% whereas Human--Object Interaction drops to 87\%, as object manipulation introduces additional reflections. Scene~4 (Fig.~\ref{fig:conf_mx}d) yields the best overall accuracy (95.28\%), with Human--Human Interaction lowest at 90\%. All scenes exceed 90\% overall; as quantified later in the ablation, phase sanitization and the dual-stream/TCN design particularly benefit high-motion classes.

\subsection{XRF55 Dataset Comparison}
To demonstrate the efficiency of the pipeline proposed in this work, we compare the results obtained after applying the WiFuse framework with those reported by Wang et al. \cite{Wang} called Deep Mutual Learning (DML). For clearer visualization, Table \ref{tab:comparativo_resnet} presents a comparison of different approaches applied to the same dataset, separated by scenarios using full DML and DML with only CSI information to ensure a fair comparison. It is worth noting that the DML-only CSI scenario uses only CSI amplitude.

\begin{table}[H]
    \centering
    \caption{Scenario-Based Accuracy Comparison Between Models}
    \label{tab:comparativo_resnet}
    \small
    \begin{tabularx}{\columnwidth}{
        >{\hsize=0.7\hsize\raggedright\arraybackslash}X 
        >{\hsize=1.3\hsize\centering\arraybackslash}X 
        >{\hsize=0.9\hsize\centering\arraybackslash}X 
        >{\hsize=1.1\hsize\centering\arraybackslash}X 
    }
        \toprule
        \textbf{Scenario} & \textbf{WiFuse*} & \textbf{DML} & \textbf{DML Only CSI} \\
        \midrule
        Scene 1 & 93.24\% & 93.66\% & 89.90\% \\
        Scene 2 & 93.45\% & 93.24\% & 89.00\% \\
        Scene 3 & 93.27\% & 95.66\% & 90.50\% \\
        Scene 4 & 95.28\% & 96.67\% & 91.30\% \\
        \bottomrule
        \multicolumn{4}{p{\dimexpr\columnwidth-2\tabcolsep}}{\footnotesize{*WiFuse denotes the architecture proposed in this work. Impact (\%) represents the relative performance gain or loss compared to the DML baseline.}}
    \end{tabularx}
\end{table}

Analyzing the proposed table and comparing our WiFuse with the DML approach used in the authors’ work, a small difference in performance can be observed. In fact, in the same Scenario 2, our method slightly outperforms the original proposal by 0.21 percentage points. It is worth noting that our entire pipeline considers only the raw CSI data available in the public dataset. We further note a protocol difference: the official XRF55 evaluation~\cite{Wang} adopts a 7:3 train--test ratio, whereas our pipeline uses an 80/20 split, and the DML figures reported here are those published in~\cite{Wang} under the official protocol. Because our partition is sample-wise, the same subjects may appear in both the training and test sets; the XRF55 results therefore characterize within-scene robustness rather than cross-subject generalization, which is assessed separately in the domain-adaptation and cross-domain experiments below.

\begin{figure}[ht!]
    \begin{center}
    \includegraphics[width =0.5\textwidth]{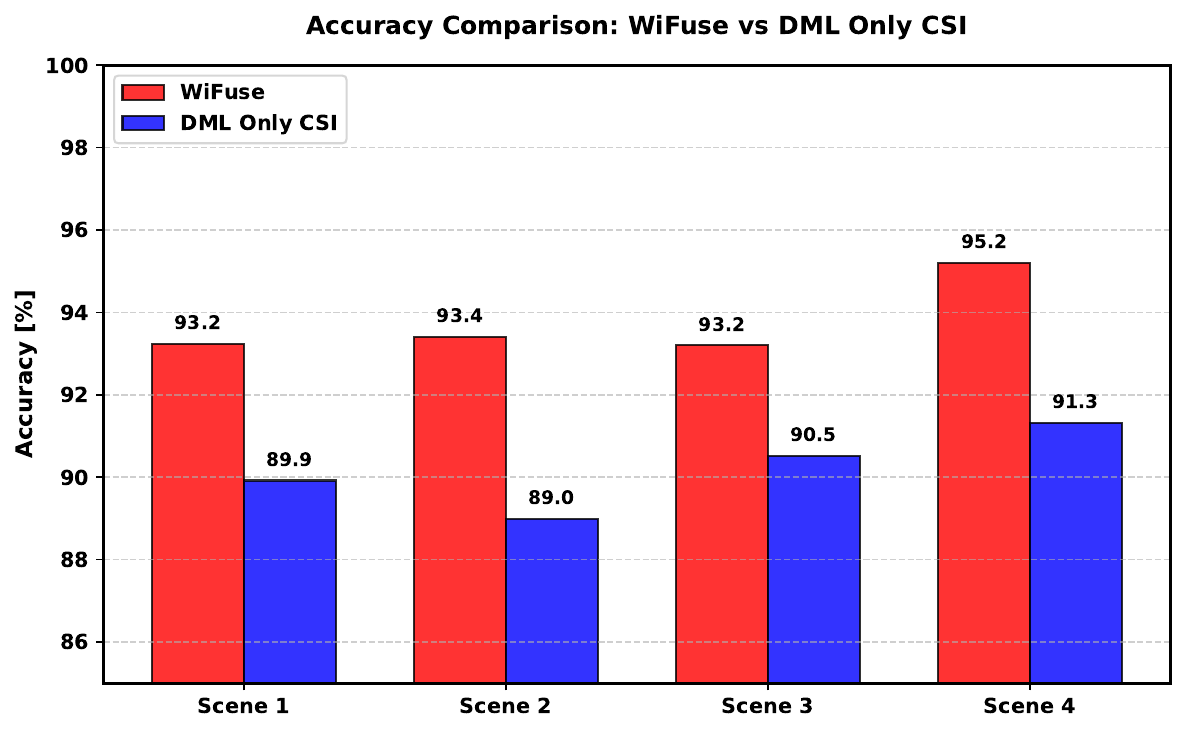}
    \caption{Classification accuracy: comparison between the proposed WiFuse and ResNet Deep Mutual Learning (DML) using only CSI.}
    \label{fig:amp_csi}
    \end{center}    
\end{figure}

A comparison that can be established, aiming to be more productive and to eliminate possible ambiguities, concerns the application of both pipelines using only CSI-based HAR data. In other words, the modalities related to RFID and mmWave are removed. The results of this comparison are presented in Fig. \ref{fig:amp_csi} for improved visualization.

In the graph presented in Fig. \ref{fig:amp_csi} it can be observed that the inclusion of our proposed pipeline and its comparison with the DML approach using only CSI data yield substantial gains. Among all evaluated scenarios, the largest performance gain is observed in Scene 2, with an improvement of 4.45 percentage points. The gains follow the environmental pattern expected from the Delay–Doppler stream: they are largest in the highly reflective Scenes 2 and 4 (4.45 and 3.98 percentage points) and smallest in the empty Scene 3 (2.77 percentage points).

Importantly, full DML benefits from three modalities (RFID, mmWave, CSI), whereas WiFuse uses CSI alone; its gains stem from phase sanitization, the dual-stream representation, and the hybrid ResNet--TCN, making it closer to real-world deployments where auxiliary modalities are unavailable. The Delay–Doppler stream improves discrimination for high-motion activities, and the TCN with transfer learning is central to the observed gains. The framework also remains effective under data scarcity, sustaining strong accuracy in the three-subject Scenes~2--4.

\subsection{Model Benchmarks}

To evaluate the efficiency of different neural networks combined with ResNet, a new experiment was conducted, considering the same preprocessing steps as well as the initial standalone ResNet application. For each experiment, the final stage of the pipeline was modified, that is, the ResNet weights were loaded into different hybrid architectures.

Since accuracy alone may lead to misleading conclusions, additional metrics were included in the experiments, namely precision, recall, and F1-score, in order to better measure the different classification performance rates. As shown in Table \ref{tab:metric_nn}, the WiFuse achieved the best overall results when compared to all other models. 

When compared to ResNet–BiGRU, which achieved the second-best performance, WiFuse obtained gains of approximately 0.46\%, 0.64\%, 0.47\%, and 0.52\% in accuracy, precision, recall, and F1-score, respectively. The performance gain ($G$) is computed using the following equation:

\begin{equation}
G = \left( \frac{V_{\text{final}} - V_{\text{initial}}}{V_{\text{initial}}} \right) \times 100
\end{equation}

where $V_{final}$ represents the metric value achieved by the best-performing model WiFuse, and $V_{initial}$ denotes the metric value obtained by the reference model ResNet–BiGRU.

The TCN outperforms the recurrent BiGRU \cite{Cho} mainly through its ability to model long-range dependencies and longer sequences without parameter inflation and at lower sequential cost. The ResNet--Transformer \cite{Trejos}, used here as a lightweight variant with frozen layers (full attention was impractical in memory), performed close to the BiGRU, while ResNet--BiLSTM and ResNet--GCN \cite{You} did not exceed 93\% accuracy. The Inception-based model \cite{Szegedy} was weakest, as its spatial-only convolutions and flatten layer poorly capture temporal dynamics at high cost. Overall, WiFuse is the most suitable choice for the evaluated datasets, though its margin is modest and other architectures may suit different settings.

\newcolumntype{C}{>{\centering\arraybackslash}X}

\begin{table}[hbt!]
\caption{Average Accuracy, Precision, Recall, and F1-Score for Hybrid Neural Networks for Scene 1}
\label{tab:metric_nn}
\centering
\setlength{\tabcolsep}{2pt} 
\renewcommand{\arraystretch}{1.2}
\begin{tabularx}{\columnwidth}{p{2.2cm} C C C C }
\toprule
\textbf{Models} & \textbf{WiFuse*} & \textbf{ResNet-BiGRU} & \textbf{ResNet-BiLSTM} & \textbf{ResNet-GCN} \\
\midrule
Accuracy (\%)   & 93.24 & 92.81 & 92.40 & 92.52 \\
Precision (\%)  & 93.58 & 92.98 & 92.59 & 92.77 \\
Recall (\%)     & 93.25 & 92.81 & 92.40 & 92.52 \\
F1 Score (\%)   & 93.23 & 92.74 & 92.33 & 92.45 \\
\bottomrule
\multicolumn{5}{@{}p{\columnwidth}@{}}{\footnotesize \textbf{Note:} This table presents a comprehensive performance benchmark of various hybrid neural network architectures, evaluated using Accuracy, Precision, Recall, and F1-score metrics. All results correspond to experimental Scene 1 of the XRF55 dataset, where WiFuse* denotes the proposed hybrid architecture.}
\end{tabularx}
\end{table}

\subsection{Ablation Tests}

\begin{figure}[ht]
    \begin{center}
      \includegraphics[width=\columnwidth]{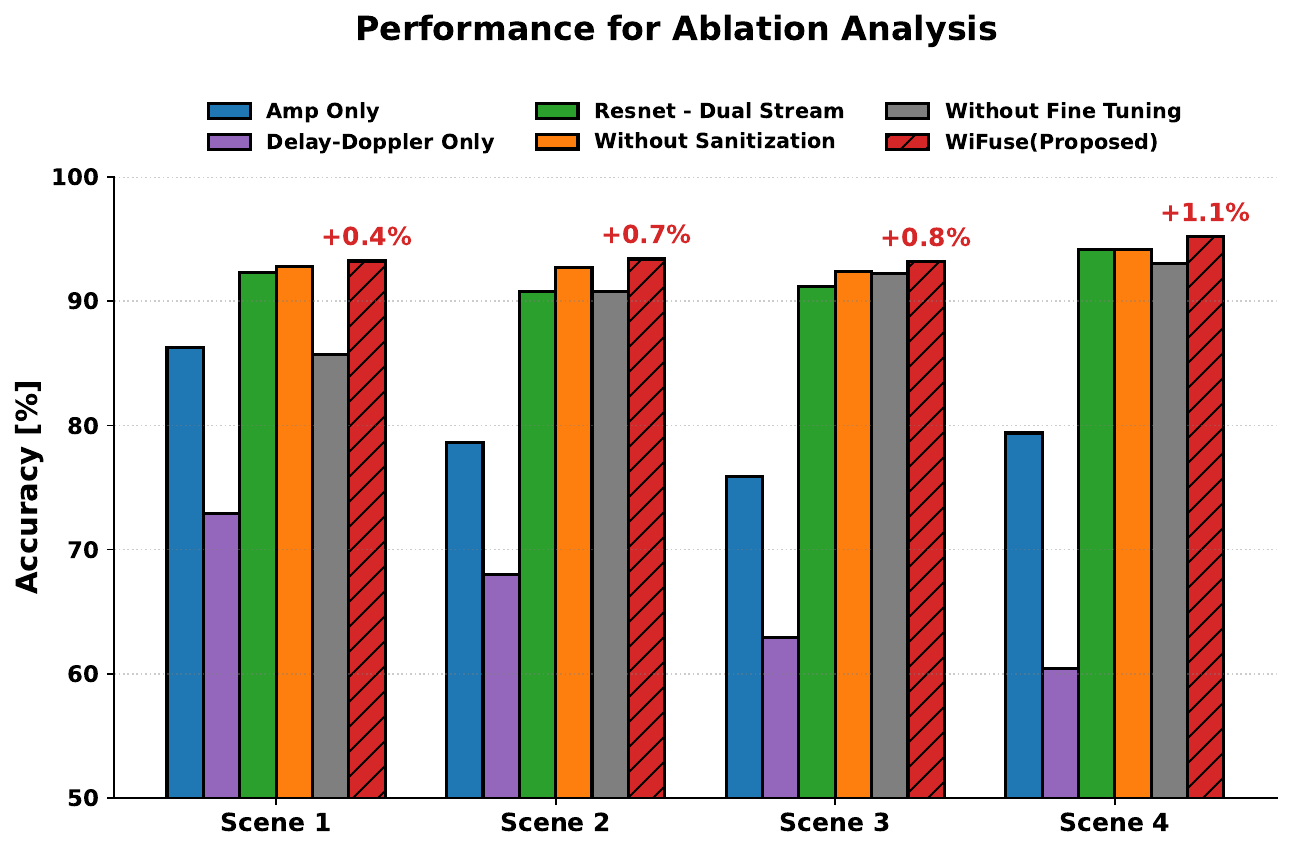}
        \caption{Comparative performance graph considering multiple experiments: amplitude-only, Delay–Doppler-only, ResNet dual-stream, ResNet–TCN without sanitization, ResNet–TCN without fine-tuning, and the proposed WiFuse in XRF55 Dataset.}
        \label{fig:ablation_}
    \end{center}    
\end{figure}

In this new experiment, several ablation processes were conducted by removing specific components of the proposed context and evaluating the resulting efficiency after each removal. In the graph shown in Fig. \ref{fig:ablation_}, it is possible to observe the classification accuracy measured under different scenarios and experimental conditions.

Initially, the following configurations and restrictions were considered:
\begin{itemize}
    \item \textbf{WiFuse:} This corresponds to the proposed pipeline, where dual-stream data related to amplitude and Delay–Doppler are concatenated and processed by a ResNet. The trained weights are then saved and subsequently reused by another hybrid neural architecture, namely ResNet–TCN;
    \item \textbf{Amplitude only:} In this experiment, only amplitude-related data were considered, while maintaining the entire ResNet structure and subsequently transferring the learned weights to the ResNet–TCN. It is worth noting that the results obtained using only amplitude differ from those reported in the original Fig. \ref{fig:amp_csi} due to differences in preprocessing. In our case, normalization and filtering steps were applied, which can significantly impact classification accuracy and were not specified in the original work;
    \item \textbf{Delay–Doppler only:} Similar to the amplitude-only experiment, only the sanitized Delay–Doppler data were considered, while preserving the same processing sequence and principles adopted in the proposed pipeline;
    \item \textbf{ResNet only:} In this configuration, dual-stream feature extraction was still applied; however, the framework execution was limited to training only the first neural network, without incorporating the TCN stage;
    \item \textbf{Without sanitization:} In this case, the entire pipeline was kept intact, except for the sanitization process. Specifically, z-score normalization per sample, linear regression, and other phase correction steps were removed, and only raw CSI data were considered;
    \item \textbf{ResNet–TCN without fine-tuning:} In this experiment, the dual-stream configuration was maintained; however, the pretrained ResNet was removed from the sequence. As a result, the ResNet–TCN model was trained from scratch, without transfer learning.
    
\end{itemize}

Comparing the results presented in Fig. \ref{fig:ablation_} the proposed WiFuse configuration achieved the best performance among all ablation experiments. Specifically, it outperformed the configuration without sanitization by approximately 0.4\%, 0.7\%, 0.85\%, and 1.1\% in Scenes 1, 2, 3, and 4, respectively. These results highlight the importance of the sanitization process, as it removes peaks, interference, and offsets, providing a significant gain in discriminating activity patterns.

Another important observation is that when amplitude and Delay–Doppler were evaluated independently, neither configuration achieved 90\% accuracy. The worst result was observed in Scenario 4 for Delay–Doppler alone, reaching only 60.40\% accuracy. This clearly demonstrates the effectiveness of the proposed approach: although each modality performs poorly when used individually, their concatenation significantly enhances the neural network’s recognition capability.

Regarding the other evaluated structures, namely ResNet-only and ResNet–TCN without fine-tuning, the results were very close, with differences on the order of approximately 1\%. This indicates that the transfer learning process was successfully applied, enabling the extraction of discriminative information that could not be effectively captured by standalone models.

Overall, the ablation analysis demonstrates that the proposed pipeline, when evaluated as an integrated context, achieves consistently strong performance, particularly in scenarios involving a large number of activity classes.

\subsection{Wi-MIR Dataset Validation}

To further validate the proposed pipeline, an additional dataset, the Wi-MIR dataset provided by Islam et al. \cite{Islam}, was employed in order to demonstrate the effectiveness of our approach. This dataset presents a higher level of difficulty compared to XRF55, as all activities are performed jointly by two individuals. When analyzing actions executed by multiple users, classification accuracy tends to decrease due to factors such as multipath interference, noise, and inconsistencies in movement patterns, which can adversely affect the recognition process. These aspects introduce an additional layer of complexity, making the task more challenging.

The Wi-MIR dataset comprises 17 distinct activity classes, with all activities performed within the same environment. Regarding the preprocessing scripts, only minor adaptations were required due to the presence of headers in the raw data collection files. The neural network architecture and the fine-tuning and transfer learning procedures remained exactly the same as those used previously. This consistency allows us to demonstrate that the proposed framework generalizes well across different datasets, despite variations in the number of participants and samples.

\begin{figure}[ht!]
    \centering

    \begin{minipage}[b]{0.24\textwidth}
        \centering
        \includegraphics[width=\textwidth]{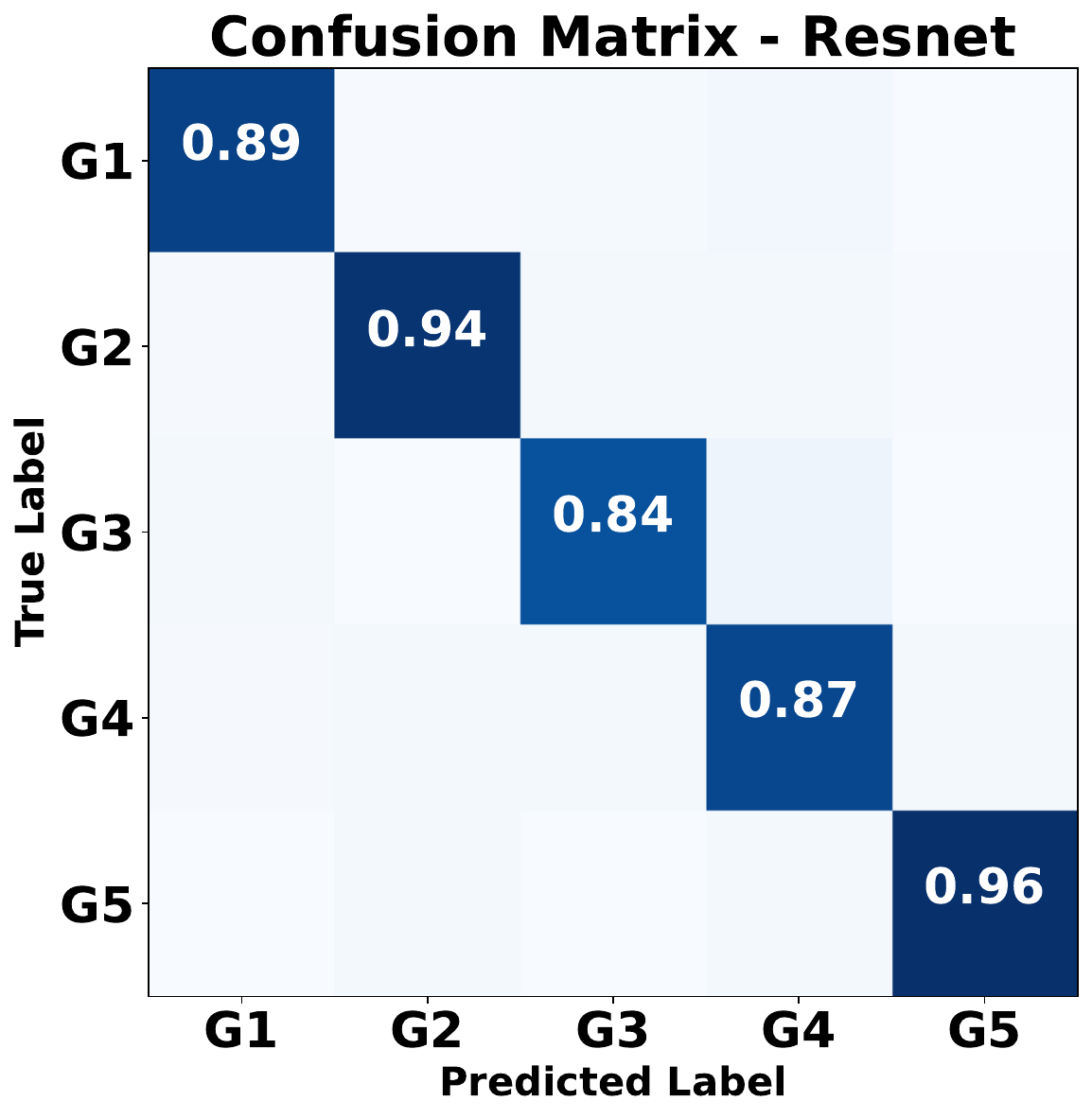}
        \text{(a) ResNet – 90.06\%}
    \end{minipage}
    \hfill
    \begin{minipage}[b]{0.24\textwidth}
        \centering
        \includegraphics[width=\textwidth]{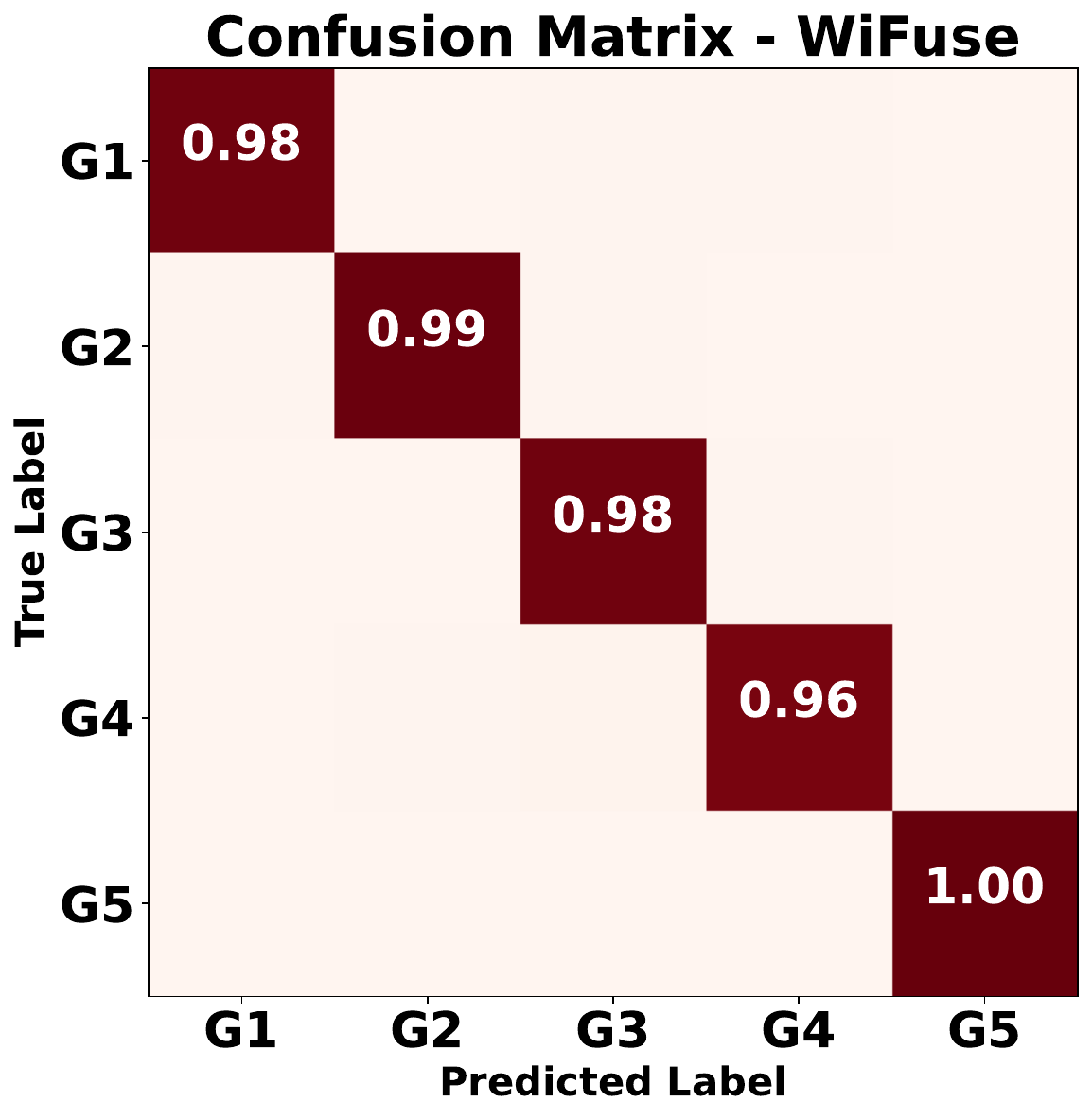}
        \text{(b) ResNet-TCN – 98.20\%}
    \end{minipage}

    \caption{Confusion matrix representations for ResNet and hybrid ResNet–TCN in the superclass context.
    Classes G1 (L1–L4), G2 (L5–L8), G3 (L9–L12), G4 (L13–L16), G5 (L17) from WiFuse.}
    \label{fig:conf_mx_wimir}
\end{figure}

Fig. \ref{fig:conf_mx_wimir} presents two confusion matrices belonging to the WiFuse pipeline: the first generated using the CSI-IRNet model, and the second obtained with the WiFuse after fine-tuning and hyperparameter optimization. Although the dataset comprises 17 distinct classes, the superclass strategy adopted in the previous experiments was employed again to improve visualization clarity. Accordingly, the activities were grouped as follows:

\begin{itemize}
    \item G1: Approaching, Bowing, Conversation, and Departing;
    \item G2: Exchanging objects, Handshaking, Helping stand up, and Helping walk;
    \item G3: Hugging, Kicking with left leg, Kicking with right leg, and Pointing with left hand;
    \item G4: Pointing with right hand, Punching with left hand, Punching with right hand, and Pushing;
    \item G5: Touching another person.
\end{itemize}

By comparing the two neural network scripts, ResNet (Fig. \ref{fig:conf_mx_wimir}a) and ResNet-TCN (Fig. \ref{fig:conf_mx_wimir}b), we observe an overall accuracy gain of approximately 8.28\%. With respect to the superclasses, a notable improvement is observed for G3, which includes activities involving close physical interaction between individuals, such as hugging, kicking, and pointing. For this group, the accuracy gain reaches approximately 6\%.

This improvement stems from transfer learning and the TCN's temporal modeling, which the standalone ResNet lacks, together with the Delay–Doppler features that suit the dataset's dynamic interactions. Among individual classes, Kicking with the right leg improves the most (about 18\%), consistent with the benefit of Delay–Doppler representations for pronounced, longer-duration motion.

\subsection{Comparison with the Wi-MIR Benchmark}

\begin{figure}[ht!]
    \centering

    \begin{minipage}[b]{0.24\textwidth}
        \centering
        \includegraphics[width=\textwidth]{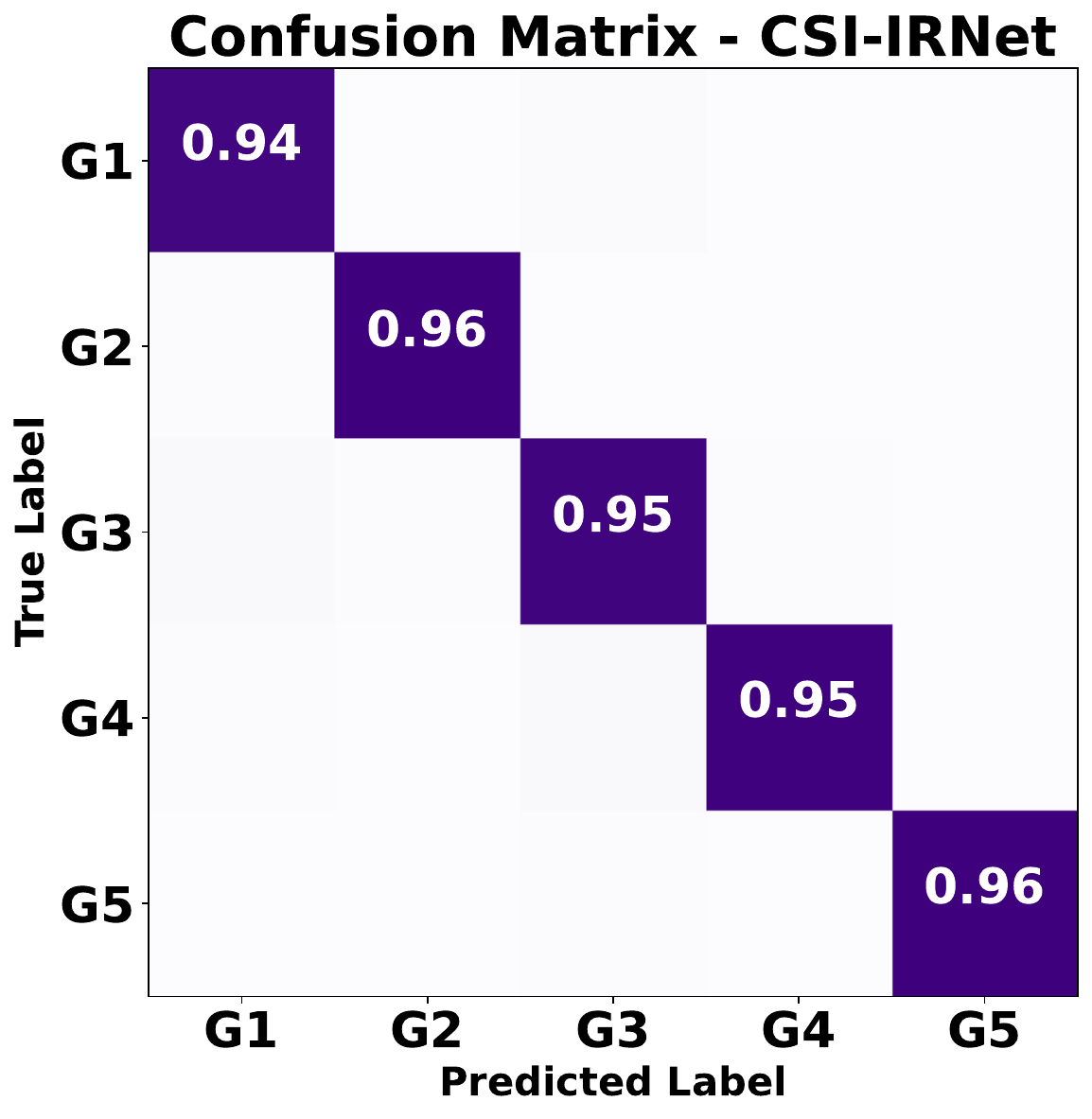}
        \text{(a) CSI-IRNet – 95.19\%}
    \end{minipage}
    \hfill
    \begin{minipage}[b]{0.24\textwidth}
        \centering
        \includegraphics[width=\textwidth]{figures/resnet_tcn_superclass.pdf}
        \text{(b) WiFuse – 98.20\%}
    \end{minipage}

    \caption{Confusion matrix representations for the original CSI-IRNet and WiFuse in the superclass context.
    Classes G1 (L1–L4), G2 (L5–L8), G3 (L9–L12), G4 (L13–L16), G5 (L17).}
    \label{fig:conf_mx_wimir_comp}
\end{figure}

Due to the unavailability of the original authors’ implementation scripts, the publicly reported results from the paper were used as a reference for comparison with this work. To ensure a fair evaluation, the same strategy was adopted, in which the original activity classes were aggregated into superclasses. The same procedure was applied when comparing the standalone CSI-IRNet with the WiFuse framework. Fig.~ \ref{fig:conf_mx_wimir_comp} presents the confusion matrices, where Fig.~\ref{fig:conf_mx_wimir_comp}a  corresponds to the reference work and Fig.~\ref{fig:conf_mx_wimir_comp}b represents the results obtained with the proposed approach.

Compared with the results reported by Islam et al. \cite{Islam} (Fig.~\ref{fig:conf_mx_wimir_comp}), WiFuse improves accuracy by 3.06\%, with consistent gains across superclasses---notable given that surpassing 90\% on this dataset is difficult---mainly due to phase sanitization and the amplitude/Delay–Doppler dual stream. The released data omit the steady-state class mentioned in the original paper, so only the available classes were scored. The $\approx$5.39\% gap between our ResNet (Fig.~\ref{fig:conf_mx_wimir}a) and the reference CSI-IRNet (Fig.~\ref{fig:conf_mx_wimir_comp}a), both nominally ResNet-based, reflects the limited reproducibility of the reference (implementation details unavailable); the decisive improvement comes from the TCN's explicit temporal modeling combined with transfer learning.

\subsection{Domain Adaptation}

We next assess generalization across datasets \cite{Singhal}. The ResNet backbone is shared and trained \emph{jointly} on both datasets, with two task-specific classification heads (55 classes for XRF55, 17 for Wi-MIR), so the feature extractor learns representations useful for both; no explicit alignment (e.g., Maximum Mean Discrepancy) is applied. In Table~\ref{tab:comparativo_wimir_xrf}, each block reports the dataset on which the jointly trained model is evaluated. Cross-environment generalization has been studied in the CSI-HAR literature, for instance through location-independent graph learning \cite{Ai}, cross-domain gesture recognition on Widar3.0 \cite{Liu}, and analyses of Doppler-feature transferability \cite{Bernaola}. Those works, however, transfer \emph{within} a single dataset (across rooms, orientations, or users); in contrast, the experiment reported here is a harder \emph{cross-dataset} setting between XRF55 and Wi-MIR, which differ in hardware, environment, sampling, and activity vocabulary.

\begin{table}[hbt!]
    \centering
    \caption{Performance Metrics Comparison: Wi-MIR vs. XRF55 for Domain Adaptation using WiFuse framework}
    \label{tab:comparativo_wimir_xrf}
    \small
    \begin{tabularx}{\columnwidth}{
        >{\hsize=0.9\hsize\raggedright\arraybackslash}X 
        >{\hsize=1.025\hsize\centering\arraybackslash}X 
        >{\hsize=1.025\hsize\centering\arraybackslash}X 
        >{\hsize=1.025\hsize\centering\arraybackslash}X 
        >{\hsize=1.025\hsize\centering\arraybackslash}X 
    }
      
        \midrule
        \multicolumn{5}{l}{\textbf{Jointly trained --- evaluated on XRF55}} \\
        \textit{Scenario} & \textit{Acc (\%)} & \textit{Prec (\%)} & \textit{Rec (\%)} & \textit{F1 (\%)} \\
        \midrule
        Scene 1 & 87.15 & 87.30 & 87.15 & 86.97 \\
        Scene 2 & 86.67 & 88.31 & 86.67 & 86.57 \\
        Scene 3 & 82.12 & 84.35 & 82.12 & 81.60 \\
        Scene 4 & 86.57 & 88.52 & 86.57 & 86.36 \\
        \midrule
       \multicolumn{5}{l}{\textbf{Jointly trained --- evaluated on Wi-MIR}} \\
        \textit{Scenario} & \textit{Acc (\%)} & \textit{Prec (\%)} & \textit{Rec (\%)} & \textit{F1 (\%)} \\
        \midrule
        Scene 1 & 49.45 & 47.82 & 49.45 & 46.29 \\
        Scene 2 & 48.64 & 48.36 & 48.64 & 47.13 \\
        Scene 3 & 48.01 & 43.64 & 48.01 & 45.14 \\
        Scene 4 & 48.18 & 46.17 & 48.18 & 46.46 \\
        \bottomrule
         \multicolumn{5}{p{\linewidth}}{\footnotesize Note: Table \ref{tab:comparativo_wimir_xrf} presents the performance metrics obtained from domain adaptation between the XRF55 and Wi-MIR datasets under different training and testing scenarios.}
    \end{tabularx}
\end{table}

Evaluated on XRF55, the jointly trained model performs well (82--87\% across scenes), indicating that Wi-MIR contributes generic, robust motion representations. Evaluated on Wi-MIR, accuracy saturates near 48--49\%, likely because residual XRF55-specific noise and hardware characteristics transfer poorly; nonetheless the metrics remain balanced, with no collapse onto a dominant class. Overall, the shared backbone learns motion patterns common to both datasets and generalizes well when the source domain is sufficiently rich.

\subsection{Cross-Domain Evaluation}

As a final stress test, we probe the limit of pure feature transfer with a zero-shot cross-\emph{dataset} protocol: the network is trained on Wi-MIR and evaluated on XRF55 with no adaptation or fine-tuning. This setting is deliberately adversarial and is reported as a diagnostic (negative) result rather than as a capability of the framework. It must be distinguished from the cross-domain protocols common in the literature, which transfer \emph{within} a single dataset---across rooms, orientations, or users, with the same hardware and label set---such as the leave-one-environment-out evaluation of Widar3.0, on which Liu et al. \cite{Liu} report $97.61\%$. Those results are therefore not directly comparable to ours: here the source and target are entirely different datasets, with unrelated capture hardware and environments and, crucially, disjoint label spaces (17 interaction classes in Wi-MIR versus 55 activities in XRF55).

Because the two label spaces do not coincide, a Wi-MIR-trained classifier cannot predict the XRF55 categories directly; accuracy is therefore measured over a shared label set obtained by aligning the two activity vocabularies, so the score reflects the amount of transferable motion structure rather than a conventional classification accuracy. Under this protocol the accuracy is expectedly low, ranging from $34.11\%$ (Scene~4) down to $8.99\%$ (Scene~1, with an F1-score of only $7.12\%$); although every scene remains above the $1.81\%$ chance level of the 55-class problem (the ``Accuracy Gain'' column of Table~\ref{tab:comparativo_wimir_xrf_cross}), the absolute values confirm that the learned representations do not transfer across datasets without adaptation. This is consistent with the severe domain shift---antenna spacing, receiver configuration, and environmental reflections all differ---and it reinforces the value of the joint-training domain adaptation of the previous subsection; closing this cross-dataset gap, e.g., through explicit domain-alignment objectives, remains an open problem.

\begin{table}[hbt!]
    \centering
    \caption{Performance Metrics Comparison: Wi-MIR vs. XRF55 for Cross-Domain with Gain Analysis  using WiFuse framework}
    \label{tab:comparativo_wimir_xrf_cross}
    \small
    \setlength{\tabcolsep}{3pt} 
    \begin{tabularx}{\columnwidth}{l *{5}{>{\centering\arraybackslash}X}}
        \toprule
        \multicolumn{6}{l}{\textbf{Train: Wi-MIR / Test: XRF55}} \\
        \midrule
        \textit{Scenario} & \textit{Acc\%} & \textit{Prec\%} & \textit{Rec\%} & \textit{F1\%} & \textit{Acc Gain} \\
        \midrule
        Scene 1 & 8.99  & 10.01  & 8.99  & 7.12  & 79.87 \\
        Scene 2 & 23.20 & 27.94  & 23.20 & 22.09 & 92.20 \\
        Scene 3 & 33.25 & 38.36  & 33.25 & 32.81 & 94.56 \\
        Scene 4 & 34.11 & 38.84  & 34.11 & 33.63 & 94.69 \\
        \bottomrule
        \addlinespace[3pt]
        \multicolumn{6}{p{\linewidth}}{\footnotesize Note: Table \ref{tab:comparativo_wimir_xrf_cross} presents the performance metrics obtained from cross-domain evaluation between the XRF55 and Wi-MIR datasets. The analysis highlights the accuracy gain achieved across various training and testing scenarios. Here, Accuracy Gain $=(\mathrm{Acc}-p_c)/\mathrm{Acc}$, where the chance level is $p_c=1/55\approx1.81\%$; it measures the fraction of predictions beyond chance and must be read together with the (low) absolute accuracy.}
    \end{tabularx}
\end{table}

\subsection{IBIS Comparison}

For comparative purposes, we evaluated the proposed WiFuse framework against our previously developed approach, named IBIS \cite{Fernandes}. The IBIS framework leverages motion dynamics and micro-Doppler variations combined with an Inception + BiLSTM neural architecture, followed by an SVM classifier to improve generalization in HAR tasks. For this comparison, its preprocessing pipeline was adapted to the dual-stream configuration while preserving its original structure.

In Table \ref{tab:comparativo_geral}, the XRF55 dataset was evaluated across Scenarios 1 to 4 under identical conditions for both frameworks. The most significant performance gain was observed in Scenario 2, where the proposed WiFuse achieved 93.45\% accuracy, compared to 90.10\% for IBIS, resulting in a gain of 3.35 percentage points.

For the remaining scenarios the gap is smaller, indicating that the TCN and attention mechanisms drive the improvement. The advantage is largest where class separation is hardest: Scenario~1 has 30 subjects and high intra-class variability, where the SVM decision boundaries of IBIS are harder to optimize \cite{HsuLin} and its BiLSTM \cite{Xinze} is prone to gradient dispersion over long sequences, whereas the three-subject Scenes~2--4 yield similar accuracy for both frameworks. Architecturally, the dilated-convolution TCN \cite{Reiter} of WiFuse processes whole temporal windows in parallel with a large receptive field and stable training, and its attention layers suppress noise-related variation, while IBIS couples an Inception--BiLSTM extractor with an SVM whose performance degrades as the number of classes and the dataset size grow. Overall, neither model is universally superior, but the dual-stream (amplitude + Delay–Doppler) design of WiFuse improves robustness in the more complex, high-class-count settings, whereas IBIS---designed for single-source Doppler/phase data---remains competitive on lower-noise data.

\begin{table*}[hbt!] 
    \centering
    \caption{Comprehensive Performance Metrics Comparison Between Models for XRF55 Dataset}
    \label{tab:comparativo_geral}
    \small 
    \begin{tabularx}{\textwidth}{ 
        >{\hsize=1.2\hsize\raggedright\arraybackslash}X 
        *{8}{>{\hsize=0.975\hsize\centering\arraybackslash}X} 
    }
        \toprule
        & \multicolumn{4}{c}{\textbf{WiFuse*}} & \multicolumn{4}{c}{\textbf{IBIS}} \\
        \cmidrule(lr){2-5} \cmidrule(lr){6-9}
        \textbf{Scenario} & \textbf{Accuracy} & \textbf{Precision} & \textbf{Recall} & \textbf{F1 Score} & \textbf{Accuracy} & \textbf{Precision} & \textbf{Recall} & \textbf{F1 Score} \\
        \midrule
        Scene 1 & 93.24 & 93.58 & 93.25 & 93.23 & 91.05 & 91.18 & 91.05 & 90.99 \\
        Scene 2 & 93.45 & 93.72 & 93.35 & 93.32 & 90.10 & 91.08 & 90.10 & 89.93 \\
        Scene 3 & 93.27 & 93.61 & 93.19 & 93.25 & 90.71 & 91.34 & 90.71 & 90.43 \\
        Scene 4 & 95.28 & 95.71 & 95.14 & 95.33 & 93.43 & 94.04 & 93.43 & 93.46 \\
        \bottomrule
        \multicolumn{9}{p{\textwidth}}{\footnotesize Note: This table presents a performance comparison between two distinct frameworks: the proposed WiFuse* and the Inception-BiLSTM with SVM post-processing (IBIS). Both architectures are evaluated across multiple metrics, including Accuracy, Precision, Recall, and F1-score.}
    \end{tabularx}
\end{table*}

To perform a reliable comparative analysis of the capabilities of WiFuse and IBIS, a table \ref{tab:comparativo_wimir_xrf_ibis} was constructed considering the domain adaptation setting, where two datasets, Wi-MIR and XRF55, were used during the training and testing stages. The results show that in Scenarios 1, 3, and 4, IBIS was able to extract more discriminative features, achieving slightly higher accuracy than WiFuse. This behavior can be largely attributed to the characteristics of the Wi-MIR dataset, which presents lower noise levels and richer information related to the activities, enabling better generalization for both frameworks.

In contrast, when XRF55 is used for training and Wi-MIR for validation, a noticeable accuracy drop of approximately 10\% is observed for IBIS. One possible explanation is that, even after the sanitization process, artifacts and residual reflections remain in the data, which negatively affects the model’s performance. Therefore, it can be observed that in scenarios with richer motion information both IBIS and WiFuse exhibit similar performance. However, when the scenario requires better class separation, WiFuse tends to outperform IBIS. It is also important to note that IBIS was adapted for this comparative analysis, since it was originally designed to operate with Doppler data from a single sensing source.

\begin{table}[hbt!]
    \centering
    \caption{Performance Metrics Comparison: Wi-MIR vs. XRF55 for Domain Adaptation using IBIS framework}
    \label{tab:comparativo_wimir_xrf_ibis}
    \small
    \begin{tabularx}{\columnwidth}{
        >{\hsize=0.9\hsize\raggedright\arraybackslash}X 
        >{\hsize=1.025\hsize\centering\arraybackslash}X 
        >{\hsize=1.025\hsize\centering\arraybackslash}X 
        >{\hsize=1.025\hsize\centering\arraybackslash}X 
        >{\hsize=1.025\hsize\centering\arraybackslash}X 
    }
      
        \toprule
        \multicolumn{5}{l}{\textbf{Jointly trained --- evaluated on XRF55}} \\
        \textit{Scenario} & \textit{Acc (\%)} & \textit{Prec (\%)} & \textit{Rec (\%)} & \textit{F1 (\%)} \\
        \midrule
        Scene 1 & 88.24 & 88.45 & 88.24 & 88.18 \\
        Scene 2 & 85.25 & 86.66 & 85.25 & 84.98 \\
        Scene 3 & 84.14 & 85.96 & 84.14 & 83.89 \\
        Scene 4 & 87.17 & 87.82 & 87.17 & 86.90 \\
        \midrule
        \multicolumn{5}{l}{\textbf{Jointly trained --- evaluated on Wi-MIR}} \\
        \textit{Scenario} & \textit{Acc (\%)} & \textit{Prec (\%)} & \textit{Rec (\%)} & \textit{F1 (\%)} \\
        \midrule
        Scene 1 & 35.73 & 36.49 & 35.73 & 36.01 \\
        Scene 2 & 34.64 & 36.33 & 34.64 & 35.33 \\
        Scene 3 & 36.82 & 37.35 & 36.82 & 37.00 \\
        Scene 4 & 36.36 & 36.70 & 36.36 & 36.45 \\
        \bottomrule
         \multicolumn{5}{p{\linewidth}}{\footnotesize Note: Table \ref{tab:comparativo_wimir_xrf_ibis} presents the performance metrics obtained from domain adaptation between the XRF55 and Wi-MIR datasets under different training and testing scenarios.}
    \end{tabularx}
\end{table}

\subsection{Computational Complexity}

In deep learning–based studies, the analysis of computational cost is highly relevant, as it reflects the hardware resources and execution time required during model operation. With this in mind, an explicit comparative analysis is presented in Table \ref{tab:complexity_analysis}, primarily comparing the proposed WiFuse framework with other networks using the XRF55 dataset.

\begin{table*}[hbt!]
\caption{Comparison of computational complexity across different architectures and frameworks}
\label{tab:complexity_analysis}
\centering
\scriptsize
\setlength{\tabcolsep}{2pt} 
\renewcommand{\arraystretch}{1.4}

\begin{tabular}{l *8{>{\centering\arraybackslash}p{2.70cm}}}
\toprule
\textbf{Metric} & \textbf{WiFuse} & \textbf{DML} & \textbf{IBIS} & \textbf{ResNet-BiGRU} & \textbf{ResNet-BiLSTM} & \textbf{ResNet-GCN} \\
\midrule
Params ($10^3$)           & 7350.97 & 20681.27 & 3769.61 & 6566.07 & 7354.55 & 4580.53  \\
Cost (GFLOPs)             & 1.82729 & 5.15090  & 0.10554 & 1.07730 & 1.13838 & 1.04210 \\
Latency (ms)              & 3.61    & 1.58     & 11.06   & 5.31    & 5.26    & 3.84         \\
Memory (GB)               & 0.0434  & 0.1048   & 0.1368  & 0.0777  & 0.0837  & 0.0321    \\
\bottomrule

\multicolumn{9}{@{}p{\textwidth}}{\footnotesize \textit{Note:} GFLOPs for ResNet and IBIS models are estimated based on specific layers (GRU, BiLSTM, GCN). Latency for IBIS includes extraction, normalization, and SVM processing.}
\end{tabular}
\end{table*}

Among the evaluated parameters is the total number of trainable parameters, which represents the number of variables learned by the neural network during training. In general, a higher number of parameters is associated with a larger and potentially more complex model. It is important to note that the original DML framework employs a shared backbone across three modalities, namely Wi-Fi, RFID, and mmWave. Consequently, when compared to our approach, which relies solely on Wi-Fi CSI, the proposed model achieves a 64.65\% reduction in the total number of parameters. Regarding the other networks, the ResNet–Inception model exhibits the lowest number of parameters. Although it presents a reduced parameter count, its performance is inferior, as observed in Table \ref{tab:comparativo_geral}, where it achieves the lowest accuracy among the evaluated approaches.

With respect to the computational cost measured in GFLOPs, the results confirm that the proposed model requires significantly fewer floating-point operations per processed sample, achieving a reduction of 64.52\% compared to the DML model. The latency metric reflects the inference time required for a single forward pass. The proposed model exhibits a higher latency than the DML model (3.61~ms versus 1.58~ms), which may affect the perceived responsiveness; this overhead can be attributed to the dilated convolutions in the TCN module, which increase the computational depth during inference.

Finally, the memory footprint, defined as the amount of VRAM required during execution, is reduced by 58.58\% compared to the DML model. Overall, the redesigned WiFuse framework achieves a substantial reduction in computational cost by operating on a single modality, demonstrating that it can maintain efficiency while lowering hardware requirements. It is important to emphasize that, although WiFuse may exhibit higher processing costs in certain comparisons with existing networks, this is justified by the accuracy achieved and its improved capability in class separation. Additionally, the IBIS framework, which incorporates elements of semantic segmentation through temporal feature discrimination, demonstrates strong performance in terms of memory usage and computational cost (GFLOPs), making it a viable alternative. Therefore, even when requiring more hardware resources in some scenarios, WiFuse proves to be more effective overall, as evidenced by the accuracy results presented in Table \ref{tab:comparativo_geral}.

\subsection{Limitations}

Although the proposed framework presents strong results for human activity recognition, it still exhibits some limitations. Even with the application of CSI phase sanitization, the framework remains dependent on the data acquisition conditions. Residual interference may still be present and can negatively affect the neural network’s recognition performance, as well as physical obstacles in the environment. The framework also performs better for activities that generate strong motion dynamics, allowing the Delay–Doppler features to be effectively mapped. In contrast, activities with low motion dynamics may not benefit as much from the Delay–Doppler representation, leading to less favorable recognition results.

An additional aspect that deserves attention concerns the number of subcarriers available in the datasets. In the datasets used in this work, the available number of subcarriers is sufficient to capture discriminative information; however, when the number of subcarriers is reduced, the lack of spectral resolution may lead to information loss and degraded performance. A further consideration is statistical robustness: the reported accuracies are single-run point estimates, and some margins over the strongest hybrid baselines, Table~\ref{tab:metric_nn}, are small relative to the expected run-to-run variation. Reporting the mean and standard deviation over multiple random seeds, together with significance testing, is left for future work.

Finally, the computational cost and processing time represent another limitation. Depending on the dataset size, both training and inference times can increase significantly, especially considering that the proposed framework operates in two stages. As a result, the overall computational cost may be higher due to the fine-tuning process involved in the hybrid architecture.

\section{Conclusion}
\label{sec:conclusao}

In this work, we proposed WiFuse, a dual-stream framework whose central novelty is the feature-level fusion of a denoised time-domain amplitude stream with a sanitized-phase Delay–Doppler stream, processed by a hybrid ResNet--TCN architecture with channel and spatio-temporal attention and trained through a decoupled two-stage transfer learning strategy. To our knowledge, this joint exploitation of temporal signal energy and motion-induced velocity dynamics within a single spatio-temporal model has not been explored in prior CSI-based HAR. The experiments confirmed that phase sanitization combined with the joint dual-domain representation yields rich, discriminative features.

Comparative evaluations against existing works and alternative hybrid neural architectures indicate that WiFuse achieves the best overall performance among the compared methods. The ablation study further revealed that the efficiency of the proposed pipeline depends on the integration of all its stages, highlighting the importance of the complete sanitization, fusion, and transfer learning process to fully exploit CSI data.

It was also observed that activity classes involving more pronounced motion patterns benefited most from the proposed architecture. Moreover, under joint-training domain adaptation the framework produced stable and balanced metrics across datasets, whereas zero-shot cross-domain transfer remained limited, consistent with the recognized difficulty of environment generalization and motivating future domain-alignment work. In particular, accuracies of up to 98.20\% were obtained on the Wi-MIR dataset in multi-user scenarios, while maintaining strong performance across different environments. Overall, the results indicate that WiFuse provides an effective solution for Wi-Fi sensing-based human activity recognition, with competitive generalization capability and classification accuracy.

For future research, several directions can be explored. One promising avenue involves studying the coexistence of multiple wireless networks operating within the same frequency spectrum and analyzing the impact of interference during CSI acquisition. Investigating advanced preprocessing strategies to mitigate such interference effects would further enhance system robustness. In addition, a re-evaluation under the official 7:3 XRF55 protocol and an explicit cross-subject evaluation are planned to further characterize generalization.

Another important direction is the evaluation of the framework in outdoor environments, where uncontrolled external interference and multipath effects may significantly affect signal propagation and activity recognition performance. Extending the framework to these more complex real-world scenarios would contribute to improving its practical applicability.

\bibliographystyle{IEEEtran} 
\bibliography{references}

\end{document}